\documentclass{article}
\usepackage[utf8]{inputenc}
\usepackage{amsfonts}
\usepackage{amsmath}
\usepackage{amssymb}
\usepackage{array}
\usepackage{amsthm}
\usepackage{subfigure}
\usepackage{mathdots}
\usepackage{algorithm, algorithmic}
\usepackage[table,x11names]{xcolor}
\usepackage{color}
\usepackage{pgf}
\usepackage{listings} 
\usepackage{hyperref}
\usepackage{makecell}
\makeatletter
\newcommand{\N}{\ensuremath{\mathbb{N}}}

\newcommand{\R}{\ensuremath{\mathbb{R}}}

\newcommand{\norm}[1]{\left\Vert #1\right\Vert}

\newcommand\py[1]{{\lstinline!#1!}}

\DeclareMathOperator*{\range}{range}

\DeclareMathOperator{\G}{\mathcal{G}}

\DeclareMathOperator*{\F}{F}

\DeclareMathOperator{\tvar}{tvar}

\DeclareMathOperator{\rank}{rank}

\newtheorem{theorem}{Theorem}[section]
\newtheorem{lemma}[theorem]{Lemma}
\newtheorem{remark}[theorem]{Remark}
\newtheorem{definition}[theorem]{Definition}
\newtheorem{example}[theorem]{Example}
\newtheorem{corollary}[theorem]{Corollary}
\newtheorem{proposition}[theorem]{Proposition}

\author{\large Anna Breger$^{a,b,c,}$\footnote{Corresponding author. Mail addresses: \{first.lastname\}@univie.ac.at}
\and \large Clemens Karner$^{a,c}$
\and \large Martin Ehler$^a$
}

\date{\small$^a${Faculty of Mathematics, University of Vienna}\\
$^b${Department of Applied Mathematics and Theoretical Physics, University of Cambridge}\\
$^c${Center of Medical Physics and Biomedical Engineering, Medical University of Vienna}
}

\begin{document}
\title{visClust: A visual clustering algorithm based on orthogonal projections}

\maketitle
\begin{abstract}
We present a novel clustering algorithm, \textit{visClust}, that is based on lower dimensional data representations and visual interpretation. Thereto, we design a transformation that allows the data to be represented by a binary integer array enabling the use of image processing methods to select a partition. Qualitative and quantitative analyses measured in accuracy and an adjusted Rand-Index show that the algorithm performs well while requiring low runtime and RAM. We compare the results to {\color{black}$6$} state-of-the-art algorithms with available code, confirming the quality of \textit{visClust} by superior performance in most experiments. Moreover, the algorithm asks for just one obligatory input parameter while allowing optimization via optional parameters. The code is made available on GitHub and straightforward to use. 
\end{abstract}
\setcounter{tocdepth}{2}

\section{Introduction}
Clustering tackles the question of how to group given data points in a meaningful way without using any references, i.e.~target data. Naturally, this task is much more challenging than methods using target data, where it is possible to use additional data information for optimization. The terminologies for clustering versus unsupervised classification differ in the literature. Here, we refer to clustering in the sense of providing a partition without any previous adjustment or training of the method, whereas an unsupervised classification algorithm may be based on a previously optimized model {\color{black}including information of fixed classes.}

We believe it is fair to say that there is no state-of-the-art clustering algorithm that is universally successful in a standardized way. Many algorithms have been proposed viewing the problem from different angles, often relying on novel theoretical results. Unfortunately many of the methods lack a user friendly implementation; often they are not made publicly available at all or lack the potential for straight forward use by non-experts. 

This might be a reason why one of the most used clustering algorithms still is \textit{k-Means}, cf. \cite{macqueen1967some}, and variants thereof (e.g. \cite{ovclust, LIKAS2003451,Zhang01generalizedk-harmonic}). The \textit{k-Means} algorithm is based on computing cluster centers and the closest neighbors regarding some distance measure, traditionally the Euclidean distance. The major advantage of \textit{k-Means} is the relatively fast runtime and clear interpretability. On the other hand the original algorithm comes with many well understood drawbacks such as inconsistency and inflexibility regarding inhomogeneous cluster sizes and densities, see e.g. \cite{altkmeans}. In terms of related approaches that are based on probability models, \textit{Gaussian mixture models (GMM)} have been well studied and implemented for clustering, see e.g. \cite{YANG20123950,gmmbook}. They allow more flexible densities, but {\color{black}are} again depending highly on a good choice of initial centers. For more complex data sets neither \textit{k-Means} nor \textit{GMM} based methods yield generally sufficient results. 

Many research fields within mathematics and computer science applied their own methods to develop novel clustering methods, for an overview of standard methods see e.g.~\cite{10.5555/1162264,JAIN2010651}. A recent popular approach to clustering are density-based algorithms, cf.~\cite{densbas}, grouping data points that are densely packed. Currently a well-known one is given by \textit{DBSCAN} \cite{Ester1996ADA}. A major drawback of \textit{DBSCAN} is the non sufficient
applicability to data with clusters of varying densities, see e.g.~\cite{dbscancrit}, which was overcome by enhanced versions, {\color{black} e.g.~\textit{OPTICS} \cite{Ankerst99optics}, 
\textit{spectACl} \cite{spectacl}, \textit{DBHD} \cite{10027741}.} It is a challenge in practice that the success of traditional density-based clustering methods is sensitive to the choice of several obligatory input parameters. Naturally, these algorithms work well for an optimized input parameter setting but have difficulties to succeed without a comprehensive parameter search within the provided ranges, requiring prior data knowledge. Therefore, in these cases, general default suggestions for the input parameters are not easy, or even possible, to provide. 

{\color{black}As deep-learning became very popular in the last decade, many methods in that domain have been introduced and successfully applied to high-dimensional data, see e.g.~the survey papers \cite{10.1093/bib/bbz170, ovclu, 10049408}. A huge drawback of deep learning based methods is the need of huge sample sets to allow proper generalization to external data \cite{10.5555/3495724.3495986,SUN2022109346}}. 

Here, we will introduce an algorithm, \textit{visClust}, that is based on the identification of dense areas by using image processing tools on transformed data representations. The algorithm asks for just one input parameter in the default setting (the number of desired clusters) and allows further optimization through optional parameters, while being easy to use. The out of the box implementation is made available on a GitHub repository \footnote{\color{black}\url{https://github.com/charmed-univie/visclust}}.

Our proposed algorithm consists of several steps that are repeated iteratively. Orthogonal projections from a chosen set are applied to identify a beneficial lower dimensional data representation that allows a visual identification of the underlying cluster structure. {\color{black}This is motivated by results on random orthogonal projections yielding beneficial lower dimensional representations}, see e.g. \cite{ortho, Achlioptas:2003wo, Thanei2017}. Via a novel approach, the new data representations are interpreted as a binary image array preserving the visual structure. This enables clusters to be subsequently selected with the help of image processing methods. To identify connected components, diffusion filtering and other sophisticated nonlinear diffusion processes for image segmentation are available, cf. \cite{weickert}. Despite the success of PDE based methods, here, we favour speed and simplicity. Therefore, we process the image by applying Gaussian filtering and thresholding to identify densely packed regions instead. The final partition is selected by comparing the output to the selected number of clusters.

We extensively evaluate our method by comparing the outcome to {\color{black}$6$} state of the art methods {\color{black}that do provide an open access code with default parameters}, namely \textit{SVM, k-Means, GMM, {\color{black} spectACl }}, the adaptive graph learning method \textit{ELM-CLR} \cite{LIU2018126}, {\color{black} and the deep clustering algorithm \textit{AdaGAE} \cite{adagae}} on several synthetic and publicly available data sets. Our evaluation includes visual comparison and quantitative comparison regarding {\color{black}accuracy (ACC),} an adjusted Rand index (ARI) as well as runtime and RAM {\color{black}for common classification experiments}. Note that for the ARI evaluation metric we use the one-sided random model for fixed number of clusters as suggested in \cite{arinew}. Our numerical experiments show that \textit{visClust} outperforms the compared methods on several data sets qualitatively and quantitatively. Moreover, it yields sufficient and stable results in the default parameter setting that asks for just one input parameter and allows improvement when the parameters are further optimized. 

The outline of the paper is as follows. In Section \ref{mathfram} the mathematical framework and theory that provides the basis for the algorithm is described. In Section \ref{alg} the algorithm is presented including implementation details. All numerical experiments and used data sets are described in Section \ref{numexp} and the results are discussed subsequently in Section \ref{disc}.

\section{Mathematical Framework}\label{mathfram}
The space of $k$-dimensional linear subspaces of $\R^d$, the so-called Grassmannian, is identified by orthogonal projections, i.e.
\begin{equation}\label{defo}
    \G_{k,d} := \{p \in \R^{d \times d}: p^2 = p, p^T = p, \rank(p) = k \}.
\end{equation}
The Grassmannian manifold is naturally equipped with an orthogonally invariant probability measure $\mu_{k,d}$ that may be viewed intuitively as the uniform distribution on $\G_{k,d}$. 
Orthogonal projections $p\in \G_{k,d}$ can be used to reduce the dimension of a high-dimensional data set  $x=\{x_i\}_{i=1}^m\subset\R^d$.
In particular, $x$ can be mapped into a lower-dimensional affine linear subspace $\bar{x}+V$, where \begin{equation}\label{samplemean}\bar{x}:=\frac{1}{m}\sum_{i=1}^m x_i\end{equation} is the sample mean and $V$ is a $k$-dimensional linear subspace of $\R^d$ with $k < d$. The lower-dimensional data representation is then given by
\begin{equation}\label{eq:11}
\{\bar{x}+p(x_i-\bar{x})\}_{i=1}^m \in \bar{x}+V, 
\end{equation}
with $\range(p)=V$. Moreover, it follows directly from \eqref{defo} that for all $p\in\G_{k,d}$ and $x_i \in x$ it holds that
\begin{equation}\label{leq}
 \|px_i\|^2 \leq  \|x_i\|^2.
\end{equation}

\begin{figure}
    \subfigure[\scriptsize Original data in $\R^3$ (left) with colored ground truth clusters (right).] {\begin{tabular*}{\textwidth}{c @{\extracolsep{\fill}}c}\includegraphics[width=0.44\textwidth]{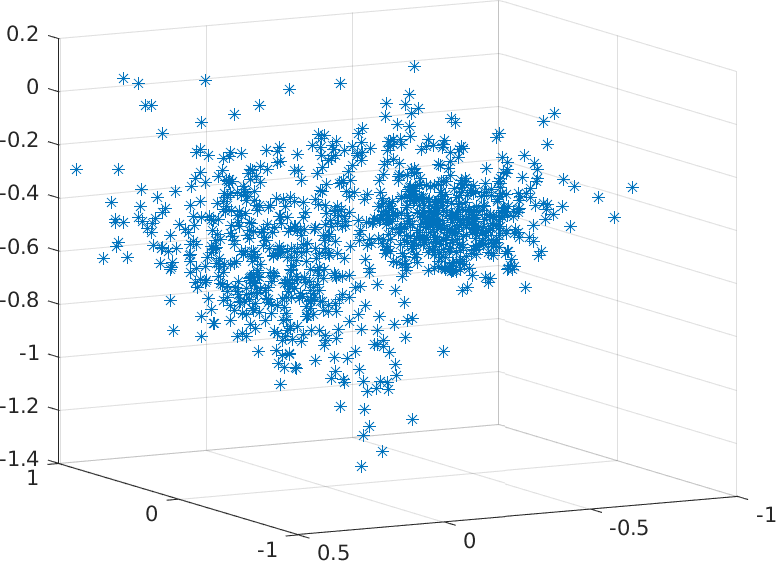}
   &\includegraphics[width=0.44\textwidth]{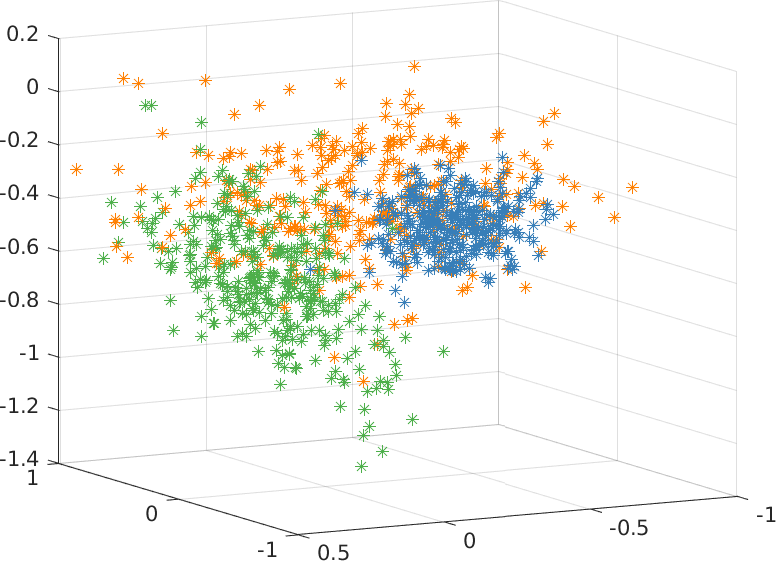}\end{tabular*}}
    \subfigure[\scriptsize Projected data in $\R^2$ (left) with colored ground truth clusters (right).]{\begin{tabular*}{\textwidth}{c @{\extracolsep{\fill}}c}\includegraphics[width=0.44\textwidth]{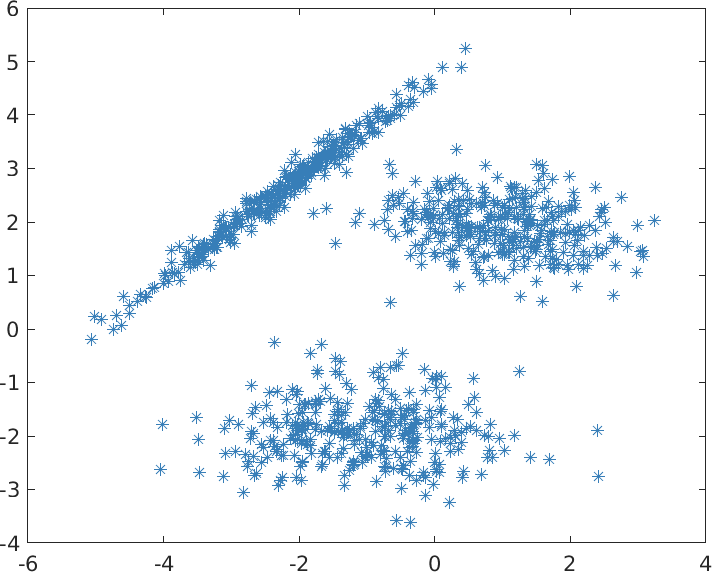} 
    &\includegraphics[width=0.44\textwidth]{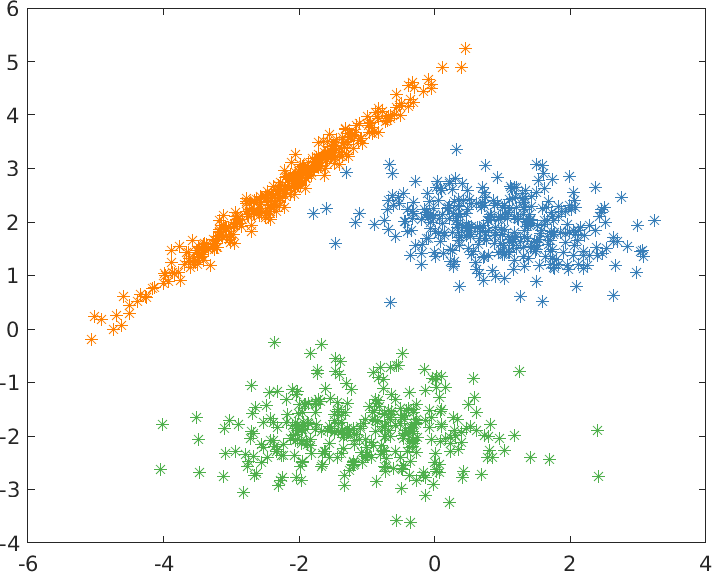} \end{tabular*}}
    \subfigure[\scriptsize Projected data using PCA in $\R^2$ (left) with colored ground truth clusters (right).]{\begin{tabular*}{\textwidth}{c @{\extracolsep{\fill}}c}\includegraphics[width=0.44\textwidth]{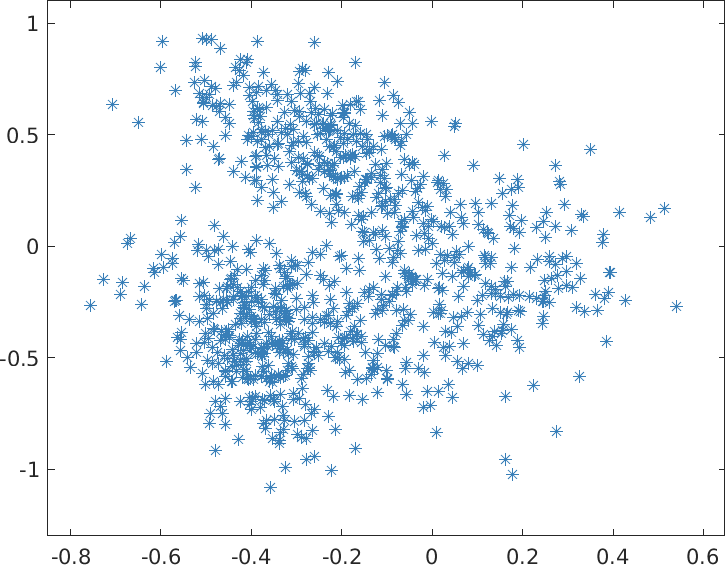} \label{subpca}
    &\includegraphics[width=0.44\textwidth]{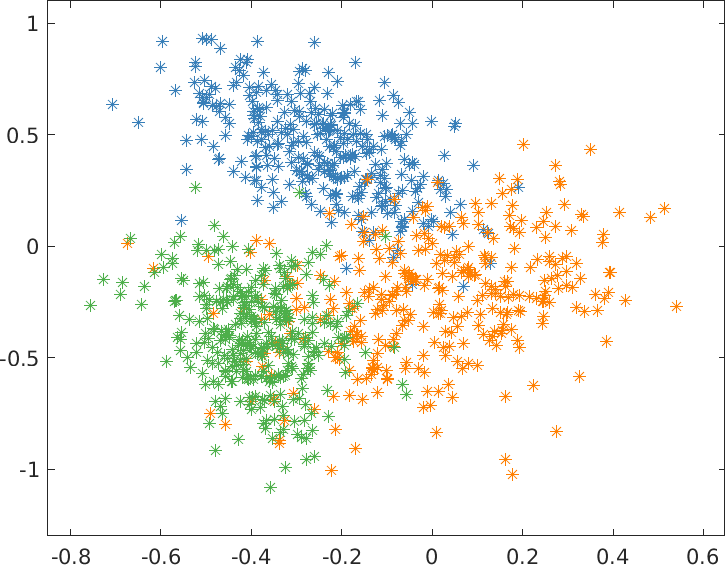} \end{tabular*}}
    \caption{\scriptsize (a) Original data $x \subset \R^3$, (b) projected data $qx \subset \R^2$ with a chosen $q \in \mathcal{V}_{2,3}$ (see Section \ref{alg}), (c) projected data using the projection provided by PCA. The synthetic dataset $x$ was generated in Python 3 using the function \textit{make\_classification} from the package \textit{sklearn}. It consists of $3$ clusters with a total of $1000$ data points. The non-overlapping clusters have a skewed normal distribution with standard deviation $1$ within a $3$-dim hypercube of side-lengths $4$. On the left the unlabeled data are shown, the right plot corresponds to the same data with added ground truth information.}
    \label{projclus}
\end{figure}

Taking a closer look at Figure \ref{projclus} for an illustration of projected pseudo-random data $x \subset \R^3$ to $\R^2$, we observe how data representations in a lower dimensional space can emphasize visual structures in given data beneficially. Note that the popular orthogonal projector computed via the principal component analysis (PCA), is an element of the Grassmannian manifold optimizing the total variance (for the definition please see Section \ref{totvar}). Although this is an important kind of information to be preserved in lower dimensional data it does not necessarily provide a beneficial representation for subsequent application of clustering methods, demonstrated in Figure \ref{subpca}. See e.g. \cite{ortho} for more details about the difference of information that is preserved by PCA versus random orthogonal projections. 

Here, we wish to view all lower dimensional representations that can be computed by orthogonal projections to find a beneficial one rather than optimizing for a specific objective. In practice, of course, we cannot view an infinite amount of projections and therefore aim for a good finite representation of the overall space of orthogonal projections. To do so, we use the so-called covering radius.

\subsection{Covering radius}
The \emph{covering radius} of a set of orthogonal projections $\{p_l\}_{l=1}^n\subset \G_{k,d}$ is denoted by
\begin{equation}\label{eq:def cov}
\varrho(\{p_l\}_{l=1}^n):=\sup_{\hat{p}\in \G_{k,d}} \min_{1\leq l\leq n} \|\hat{p}-p_l\|_{\F},
\end{equation}
where $\|\cdot\|_{\F}$ is the Frobenius norm. It indicates how well the finite set of projectors represents the overall space, i.e.~if $\G_{k,d}$ is covered properly. The smaller the covering radius, the better the finite set of projections $\{p_l\}_{l=1}^n$ represents the entire space $\G_{k,d}$: It yields smaller holes and the points are better spread. 

\begin{theorem}[Reznikov and Saff \cite{10.1093/imrn/rnv342}]\label{randproj}
The expectation of the covering radius $\rho$ of $n$ random points $\{p_j\}_{j=1}^n$, i.i.d. with respect to $\mu_{k,d}$ satisfies
\begin{equation*}\label{eq:random asympt}
\mathbb{E}\rho \asymp n^{-\frac{1}{k(d-k)}} \log(n)^{\frac{1}{k(d-k)}}.
\end{equation*}
\end{theorem}

Following the definition of asymptotically optimal covering in \cite{BREGER20181}, random projections cover asymptotically optimal up to the logarithmic factor. Since the deterministic construction of the asymptotically optimal covering sequences relies on the specific choice of $k$ and $d$, we work with random projections distributed according to $\mu_{k,d}$ which allows us flexibility and fast computation in all dimensions of $\G_{k,d}$, i.e. the choice of $k$ and $d$. 

\subsection{{\color{black} Sample variance}}\label{totvar}
{\color{black} The unbiased sample variance, here referred to as \textit{total variance}} 
$\tvar(x)$ of $x=\{x_i\}_{i=1}^m\subset\R^d$, is the sum of the corrected variances along each dimension, and it holds that
\begin{equation}\label{eq:total variance}
\tvar(x) :=\frac{1}{m-1}\sum_{i=1}^m \|x_i-\bar{x}\|^2 = \frac{1}{m(m-1)} \sum_{i<j} \norm{x_i - x_j}^2,
\end{equation}
where $\bar{x}$ corresponds to the sample mean \eqref{samplemean}. {\color{black}The computation may be found in the Supplementary Material.}
Note that the total variance of $px=\{px_i\}_{i=1}^m$ coincides with the one of \eqref{eq:11} and satisfies
\begin{equation}\label{inequ}
\tvar(px)\leq\tvar(x)
\end{equation}
for all $p \in \G_{k,d}$. This follows directly from \ref{leq}.
\begin{remark}
The above is an important property for our algorithm, since \eqref{inequ} ensures that the projected data points are controllable and do not expand in an unbounded manner.  
\end{remark}

\begin{remark}
The famous Johnson-Lindenstrauss Lemma, cf. \cite{Dasgupta:2003fk}, states that for specific $d$ and $k$ the relative pair-wise distances are preserved with high probability in the lower dimensional representation obtained by a random orthogonal projection. This is a very important result, enabling the preservation of that kind of information while allowing real-time dimensionality reduction. Nevertheless, it has to be noticed that this result does not hold for smaller dimensions $k$. For our algorithm it is beneficial when the Lemma does not hold with high probability since we aim to obtain various new data representations that do not necessarily preserve the original pairwise distances. Therefore, we decided to focus on $k=2$ and $k=3$.
\end{remark}

\section{Algorithm}\label{alg}
{ \color{black}In practice the dimension reduction takes place by applying $q\in \mathcal{V}_{k,d}$ with $q^\top q = p\in\G_{k,d}$, where 
\begin{equation*}
\mathcal{V}_{k,d}:=\{ q\in\R^{k\times d} : qq^\top = I_k\}
\end{equation*}
denotes the Stiefel manifold. When taking norms, $p$ and $q$ are interchangeable, i.e., $\|q(x)\|^2 = \|p(x)\|^2$, for all $x\in\R^d$. Therefore we can use w.l.o.g. theory developed for projections in $\G_{k,d}$ although we will apply $q$ eventually.  

The random projector $p$ is generated by applying a qr-decomposition $M=qr$ to a $k\times d$ matrix M of independent standard normal distributed entries. The $k\times d$ matrix $q$ is a random matrix in $\mathcal{V}_{k,d}$, which may intuitively be called uniformly distributed. The matrix $p=q^\top q$ is distributed in $\mathcal{G}_{k,d}$ according to $\mu_{k,d}$, which fits the random projectors addressed in Theorem \ref{randproj}. While $p$ only depends on the corresponding subspace, the matrix $q$ refers to a specific choice of a basis for this subspace. The default of the algorithm computes a set of such random projections. 

In the following we will denote a set of projections on the Stiefel manifold by $p$. All implementations were done in Matlab2020b and the code is publicly available on the GitHub repository\footnote{\color{black}\url{https://github.com/charmed-univie/visclust}}. 
}

\begin{algorithm}
{\color{black}
\footnotesize{
\caption{The visClust algorithm.}\label{alg:vis}
  \begin{algorithmic}
  \STATE  \textbf{Required Input:} Data $x=\{x_i\}_{i=1}^m\subset\R^d$, number of clusters $n_c \in \N$.
  \STATE  \textbf{Optional Input:} Threshold $t \in \R$, cluster division $\eta \in [0,1]^{n_c}$, projections $p\in \R^{k \times d}$, scaling factor $s \in \R$, subsampling $u \in \N$.
  \STATE \textbf{Output:} Clustering $c \in \R^{m}$.
  \FOR{$l \in \{1, \ldots,\#p$\}}
  \STATE \textbf{Step 1:} Project data $x$ with projection $p_l$ to $\R^k$.
  \STATE \textbf{Step 2:} Quantize projected data $p_lx \in \R^k$ using \eqref{transf} to create binary 'image' $M \in \{0,1\}^\Omega$. 
  \STATE \textbf{Step 3:} Estimate standard deviation for Gaussian filter using $s$.
  \STATE \textbf{Step 4:} Apply Gaussian filter as in \eqref{gauss} to $M$ and threshold $M$.
  \STATE \textbf{Step 5:} Identify connected components for whole data set, see Section \ref{cc} 
  \STATE \textbf{Step 6:} \IF {$n_{cc} = n_c$ and $\norm{ \eta_{cc} - \eta_c}_1 < t$}
  \RETURN Clustering $cc \in \R^m$.
  \ENDIF
  \ENDFOR
  \end{algorithmic}
  } }
\end{algorithm}

\subsection{Details of substeps}\label{substeps}
\begin{figure}
    \subfigure[\scriptsize Substep 1: Data is projected with a projection from the Stiefel manifold.] {
   \includegraphics[width=0.47\textwidth]{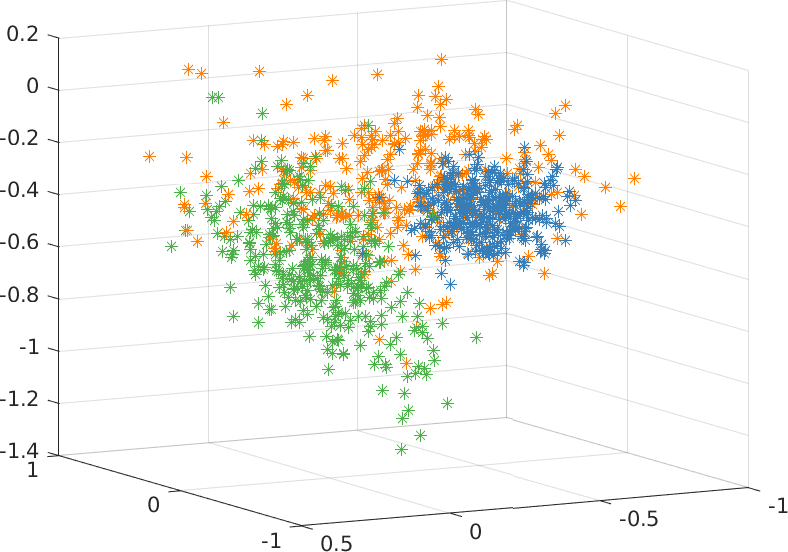} 
    \includegraphics[width=0.47\textwidth]{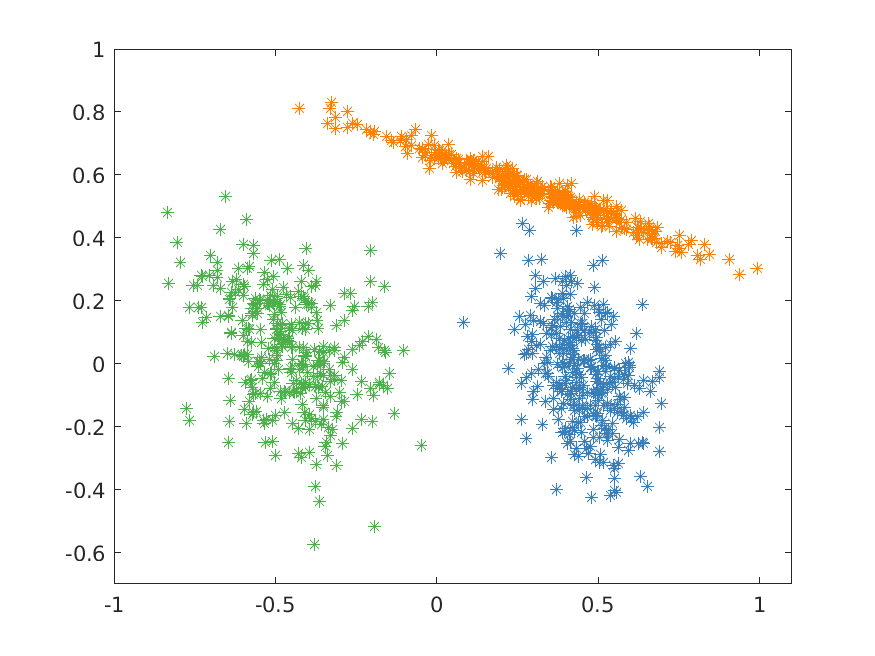} 
}
    \subfigure[\scriptsize Substeps 2 - 4: Data is interpreted as an image and a Gaussian filter is applied.] {
   \includegraphics[width=0.47\textwidth]{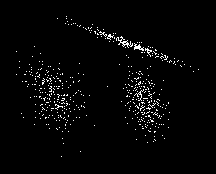}
    \includegraphics[width=0.47\textwidth]{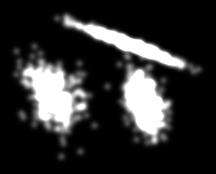} }
    \subfigure[\scriptsize Substep 5: Outliers are removed and connected components identified.] {
   \includegraphics[width=0.47\textwidth]{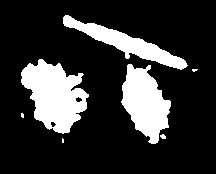} 
    \includegraphics[width=0.47\textwidth]{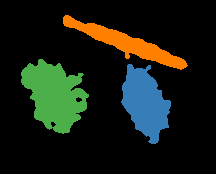} }
    \caption{\scriptsize Substeps of the algorithm described in detail in Section \ref{substeps}.}
    \label{substepsfig}
\end{figure}

{\color{black}In the following the substeps of the pseudo-code of algorithm \ref{alg:vis} are explained in detail for a fixed iteration $l$, a given data set $x=\{x_i\}_{i=1}^m\subset\R^d$ and a set of projections $p = \{p_l\}_{l = 1}^n \subset \mathcal{V}_{k,d}$. Moreover, in Figure \ref{substepsfig} all substeps are visualized.

When substep $6$ yields a sufficient evaluation in the $l$-th iteration, the algorithm terminates for $l<n$ iterations. Since the number of maximally used orthogonal projections, i.e. $n$, is fixed beforehand, it is possible to provide the user an estimated maximal runtime (corresponding to $n$ iterations) when starting the algorithm.} 
\subsubsection{Step 1 - Project data}
In the first step the provided data $x \subset \R^{d}$ is scaled between -$1$ and $1$ in each component and then linearly transformed to $\R^{k}$ with a projection $p_l$ from the provided set $p$. It follows directly that $p_l x_i \in \lbrack -\sqrt{d},\sqrt{d} \rbrack^k$ for all data points $x_i \in x$.

\subsubsection{Step 2 - Interpret projected data as image}
To allow methods from image processing, we aim to work with the data interpreted as a binary 'image' array, rather than points in a coordinate system. To do so, rescaling and data-dependent quantization is needed to first obtain an integer array. We chose to design a transformation of the lower dimensional data that allows points that differ up to two decimals to be represented by a different image pixel. If the data set contains points that are more similar, they will be represented by the same pixel. This is beneficial since very close points are assumed to be part of the same cluster. 

In particular, the projected data points $p_l x_i$ are rescaled and quantized in each dimension $r = 1, \dots, k$ by
\begin{equation}\label{transf}
z_i^{(r)} = \lfloor 100\cdot(p_l x_i ^{(r)} - \min_{j = 1, \dots m} p_l x_j ^{(r)}) \rfloor.
\end{equation}
We obtain new data points $z = \{z_i\}_{i=1}^m$ with $z_i \in \{ 0, 1, \dots, \lfloor 200 \cdot \sqrt{d} \rfloor \}^k$. The developed representation \eqref{transf} allows to preserve the visual structure of the points in coordinate representation, see Figure \ref{substepsfig} (a) and (b). The achieved image size varies depending on the minimum and maximum value of the projected data $p_l x$.

Note that the transformed data points only consist of positive integer values. We then receive the binary image array $B \in \{0,1\}^{\Omega}$ directly by setting all indices to $1$ that equal the values in $z$ and all other indices to $0$. The image array size is given by $\Omega = \{0, \dots, \max_i z_i^{(1)}\}\times \dots \times \{0, \dots, \max_i z_i^{(k)}\}$.

\subsubsection{Step 3 - Estimate standard deviation for a Gaussian filter}
Next, we compute the standard deviation for a Gaussian filter of the form 
\begin{equation}\label{gauss}
    h_l(x) := \frac{1}{2 \pi \sigma_l^2} \exp^{\big(-\tfrac{\norm{x}}{2 s \sigma_l^2}\big)}.
\end{equation}
The success of the algorithm is sensitive to the standard deviation $\sigma_l \in \R^+$ and the size of the filter. Therefore, they are set to depend on the projected data in every iteration step $l$. If the data set consists of more than $500$ data points a subset of $500$ points is chosen randomly. From this sub-sampled data set the median of the $1000$ shortest distances within the projected data $p_lx$ is computed and scaled by the dimensions as the base value for $\sigma_l$. To enable more flexibility, we allow weighting of $\sigma_l$ by a scaling factor $s \in \R^{+}$.

\subsubsection{Step 4 - Apply Gaussian filter and threshold to image}
Subsequently we apply a discretized Gaussian filter to the binary image array $B$ by convolution. It is important here that for the identification of densely packed areas the used filter size is flexible and depends on $\sigma_l$. The obtained grayscale image is then thresholded by its resulting mean, i.e. all entries that are above the mean are preserved as $1$, everything else is set to $0$. This process is simple and fast and may well be viewed as a simplified diffusion filtering technique.

\subsubsection{Step 5 \& 6 - Identify and check number of connected components in the thresholded image}\label{cc}
First, we are discarding outliers, i.e. clusters that are smaller or equal than the Gaussian filter size. The remaining connected components are identified via a standard algorithm based on equivalence tables (see e.g. \cite{3918}). {\color{black} Each data point is then assigned to the connected component it is part of, yielding a vector $cc \in \{1, \dots, n_{cc}\}^m$ where $n_{cc}$ denotes the number of connected components. Finally, we inspect if the identified numbers of connected components, i.e. clusters, coincides with the number that was provided. In this case, we assign the previously discarded outliers to the nearest clusters and compute the deviation of the desired and the output cluster division. The cluster division describes the relative size of the clusters and is given for the desired division by a vector $\eta_c = (\eta_1, \dots, \eta_{n_c})$ with all entries in $[0,1]$ such that $\sum_{i=1}^{n_c} \eta_i = 1$. If $\norm{\eta_{cc} - \eta_c}_1$, where $\eta_{cc}$ corresponds to the output division, is smaller than the threshold $t \in \R^+$, the algorithm succeeded and the final partition is returned to the user. }

\subsection{Input parameters}\label{inpar}
In the following, the required and optional input parameters are listed. If not provided by the user, the optional parameters are set to their default values. 
\medskip \\
\textbf{Required input parameters (1)}: \\ \hspace{0.1cm} Number of clusters $n_c$ \medskip \\
\textbf{Optional input parameters (5):} 

\begin{itemize}
\item Threshold: \indent $t \in [0.01, 0.3]$ \indent \  (default: $t=0.1$) 
\item Scaling factor: \indent $s \in [0.5, 2]$ \indent \ (default: $s=1.25$)
\item Subsampling: \indent $u \in \{1, \dots, m\}$ \indent \ (default: $u = m$)
\item Dimensionality reduction: \indent $p \in \R^{k \times d}$ {\color{black}or t-SNE} \indent \ (default: random projections $p \subset \mathcal{V}_{k,d}$)
\item Cluster division: \indent $\eta \in [0,1]^{n_c}$ s.t. $\sum_{i} \eta_i = 1$ \indent \ \big(default: $\eta = (\frac{1}{n_c}, \dots, \frac{1}{n_c})$\big)

\end{itemize}

\section{Numerical Experiments}\label{numexp}
First, to allow comprehensive analyses, we use synthetic data that can be easily manipulated regarding desired properties, e.g. shapes or sizes of the clusters. Secondly, we employ publicly available data sets that also provide target values for the objective evaluation of the methods. 

All experiments {\color{black} except for the \textit{AdaGAE} algorithm} have been done using a MacBook Pro equipped with an Intel i7-4770HQ with 4 cores clocked at 2.2 GHz. {\color{black} In the case of \textit{AdaGAE}, we employed a computer cluster equipped with an AMD EPYC 7713 with 64 cores clocked at 2 GHz and an NVIDIA A100 GPU.}

\paragraph{Overview algorithms}\label{compalg}
For comparison we chose six well known, state-of-the-art classification and clustering algorithms that provide open source implementations. One of them is a supervised learning method, the famous support vector machine algorithm (\textit{SVM}), whereas the clustering algorithms are unsupervised. Therefore the results of the method have to be understood as an optimized setting in contrast to the unsupervised methods which use the available target data solely for the subsequent evaluation. The following algorithms are considered:
\begin{enumerate}
    \item[$-$] SVM (cf. \cite{Cortes1995}): Separates data points using hyperplanes that are computed based on provided target data. We apply the implementation \textit{fitcecoc} from Matlab using a Gaussian kernel.
    \item[$-$] k-Means (cf. \cite{macqueen1967some}): Groups data points to the nearest cluster center regarding some predefined distance measure. It tends to produce clusters that are equally sized and have the same spatial extent. We use the implementation \textit{kmeans} from Matlab with Euclidean distances.
    \item[$-$] GMM (cf. \cite{McLachlanGeoffreyJ2000}): Works similar as \textit{k-Means}, but fits skewed Gaussian distributions upon the data points, allowing diverse shapes. Moreover, \textit{GMM} provides probabilities for each class. We use \textit{fitgmdist} from Matlab.
    \item[$-$]{\color{black}  SpectACl (cf. \cite{spectacl}): Is a density-based clustering algorithm. It combines the benefits of the Spectral Clustering and density-based DBSCAN algorithm, while eliminating some of their drawbacks. We use the Python 3 implementation provided on Gitbucket\footnote{\color{black}\url{https://bitbucket.org/Sibylse/spectacl/src/master/}} by the authors of the paper.  }
    
    \item[$-$] ELM-CLR (cf. \cite{LIU2018126}): Is based on a feedforward neural network trained within a single step, drastically decreasing the speed in comparison to methods that require training with multiple iterations. We use the Matlab implementation provided on GitHub\footnote{\color{black}\url{https://github.com/liut0012/ELM-CLR}} by the authors of the paper.

    {\color{black}\item[$-$]AdaGAE (cf. \cite{adagae}): Is a deep clustering algorithm extending deep graph clustering to data sets without graph structure using an adaptive graph auto-encoder. We use the Python 3 implementation provided on Github\footnote{\color{black}\url{https://github.com/hyzhang98/AdaGAE}} by the authors of the paper.}
\end{enumerate}

\begin{table}
\centering
\resizebox{0.8\columnwidth}{!}{
	\begin{tabular}{ |c|c|c|c|c|c|c|c| }

		\hline
\textbf{Algorithm}& \textbf{SVM} & \textbf{k-Means} & \textbf{GMM} & {\color{black}\textbf{SpectACl} }& \textbf{ELM-CLR} & \textbf{\color{black}AdaGAE}& \textbf{visClust}\\
		\hline
		\hline
		Required& 1 & 1 & 1 & 1 &7&{\color{black}1}& 1 \\
		\hline
		Optional& 14 & 6 & 8  & 5& 0 &{\color{black}7}& 5 \\
\hline
\end{tabular} }
\caption{\scriptsize Number of required and optional input parameters for the chosen algorithms {\color{black}as required in the provided implementations. Note that all tested clustering methods require the number of clusters as an input parameter. }}
\label{paramalg}
\end{table}

The described methods naturally differ regarding their required and optional parameters, see Table \ref{paramalg}. To allow fair comparison we assume the natural scenario that the user of the methods does not have any information about internal optimal parameter choices. Therefore, we computed the results in the suggested default setting of the input parameters for all methods. The number of wanted clusters, which is here required in all tested methods, is assumed to be known. 

\begin{figure}
\centering
    {\includegraphics[width=1\textwidth]{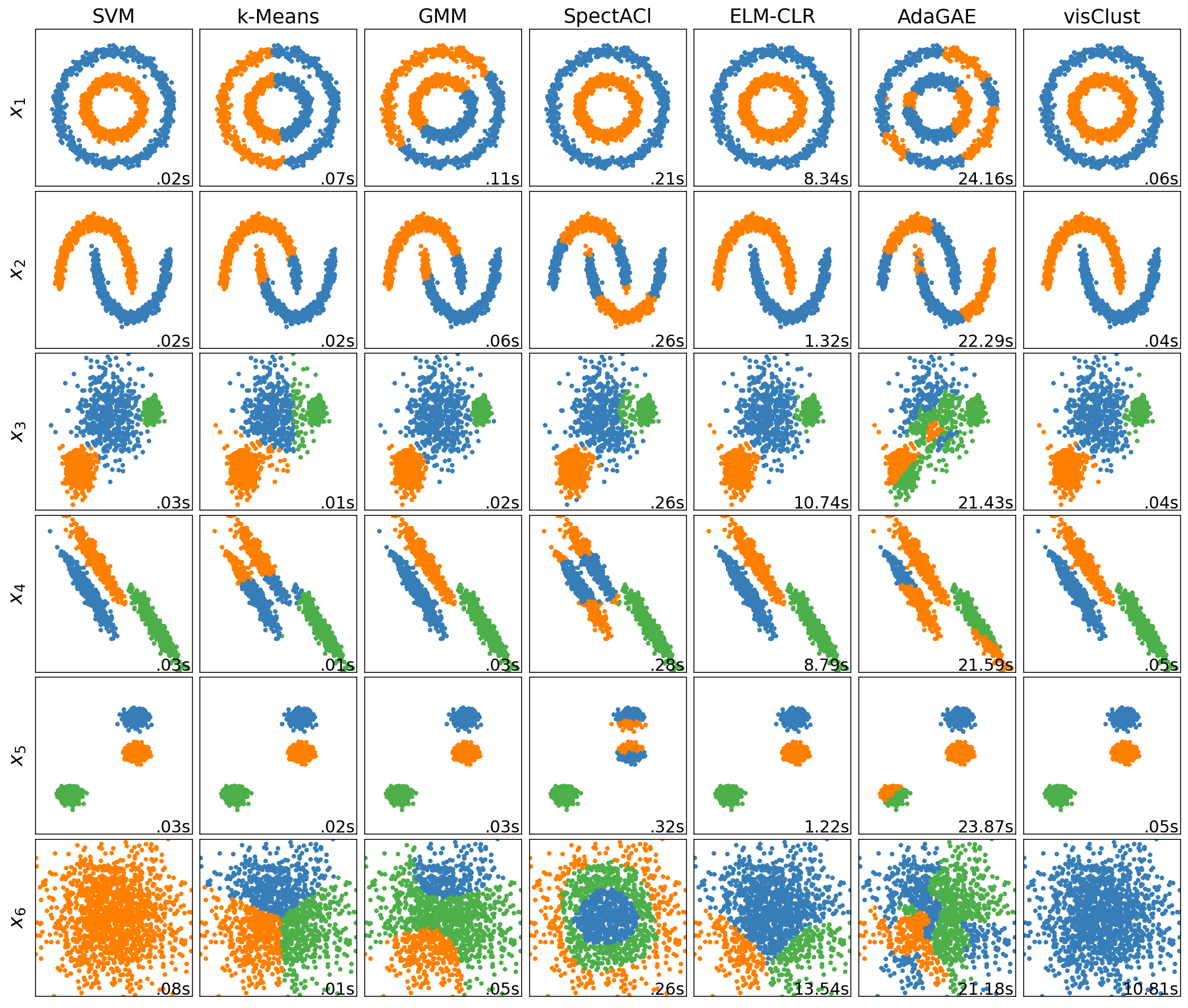}}
    \caption{\scriptsize Visual comparison of the results from the chosen clustering algorithms on $6$ synthetic data sets in $\R^2$. The needed time for termination is stated in seconds.The data sets of dimension $d=2$ consist each of $m=1500$ data points. They are generated by using Python 3 with the following functions from the module \py{sklearn}: \textit{make\_circles }(1), \textit{make\_moons} (2) and \textit{make\_blobs} (3-5). The last data set comprises of a single Gaussian cluster generated with the function \textit{random.normal} from the module \py{numpy}. The data points are randomly distributed along the different shapes with Gaussian distributions of varying standard deviations. In the last row the data consists of one big cluster but the methods were asked to return $3$ clusters. {\color{black}Note that \textit{AdaGAE} had to be computed on a different computer and therefore the stated runtime here may not be directly comparable, see Section \ref{numexp}.}} 
    \label{visComp}
\end{figure}

\subsection{Visual comparison}
First, we provide some straightforward visual experiments to compare the algorithms described in the previous section. 
In Figure \ref{visComp} we observe important clustering characteristics on $6$ synthetic data sets exhibiting different visual data structures. 
\begin{remark}
If $k=d$, the projections in the Stiefel manifold $\mathcal{V}_{k,d}$ correspond to rotations and reflections, i.e. \textit{visClust} just relies on rotated and reflected data in the previous experiments. In order to make sure that this special case is not controlling the outcome of the investigation, we also conducted equivalent experiments for $p \subset \mathcal{V}_{2,3}$ which approved the results by equivalent outcomes. 
\end{remark}

\subsection{Quantitative evaluation}
Our quantitative evaluation consists of comprehensive experiments with synthetic and real-world data. Moreover, we provide analyses of parameter choices within \textit{visClust}. As comparison metrics we use {\color{black}accuracy (ACC) and} an adjusted Rand index (ARI), described in the following, as well as the runtime measured in seconds.

\paragraph{Adjusted Rand Index (ARI)}
The Rand index \cite{ri} was proposed in 1971 to objectively compare the similarity of two partitions via pairwise comparison of the elements. The index is criticized for not correcting according to grouping elements by chance, i.e. not assessing similarities in the context of a random ensemble of partitions. Therefore an adjusted Rand index was proposed in \cite{ari}. This index is nowadays a commonly used metric to objectively asses new clustering algorithms. Interestingly, this correction relies on the assumption for the random partition ensemble to be a permutation model in which the number and sizes of clusters are fixed. This does not hold true for clustering algorithms that ask for a fixed number of clusters but allow different cluster sizes as an output. Actually, this is the case for most clustering methods, including all the methods described in Section \ref{numexp}. Therefore, in these cases the evaluation with the metric is not correct, resulting in wrong evaluation values being reported in the literature. Other random models for correction of the Rand index were suggested in 2017 \cite{arinew}, allowing different assumptions and studying the importance of a correct choice of random model. Here we use the random model that fits our evaluated algorithms, i.e. the one-sided random model for fixed number of clusters which allows different cluster size outputs therein. Despite its different random model we will also refer to it as 'adjusted Rand index (ARI)' in the following.

\subsubsection{Experiments with synthetic test data sets}\label{datas}
We create synthetic random data sets that allow straightforward manipulation for different settings enabling comprehensive runtime analyses. The basis are uniformly or normally distributed disjoint clusters. We then vary the random data $x=\{x_i\}_{i=1}^m \in \R^d$ with $n_c$ clusters according to the number of data points $m$, dimensions $d$,
and clusters $n_c$. 

\begin{figure}
\centering
 \begin{tabular}[c]{cc}
   \subfigure{
   \centering
       \includegraphics[width=0.48\textwidth]{ 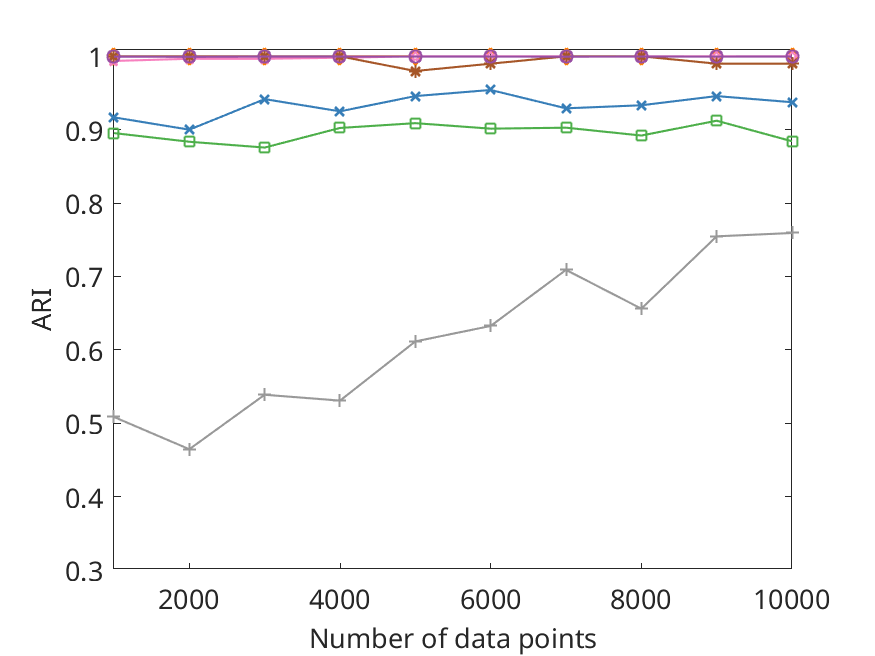} 
 }
  &
     \subfigure{
        \centering
    \includegraphics[width=0.48\textwidth]{ 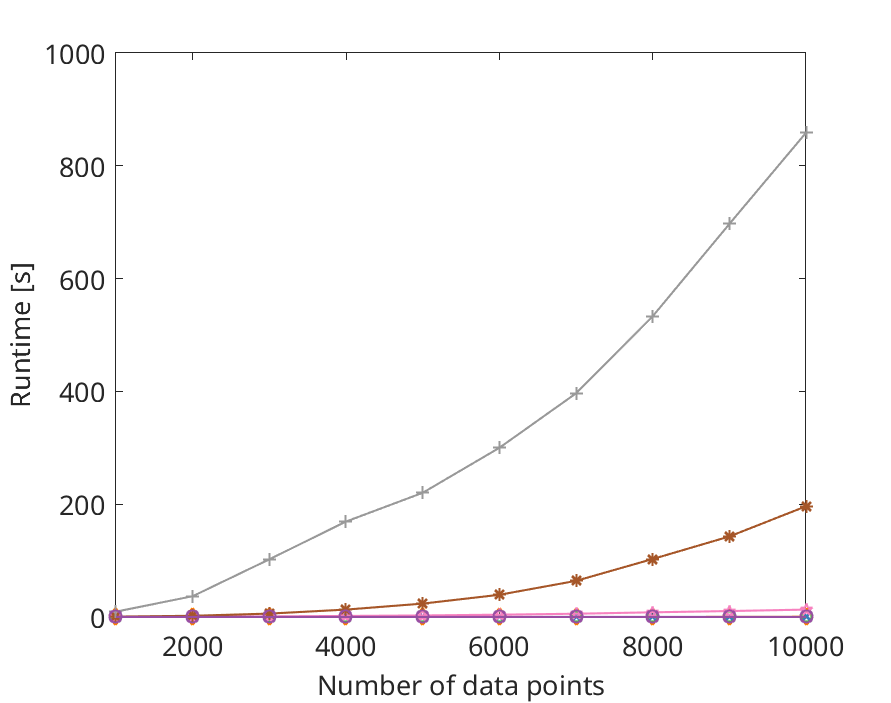} 
}
        \\
     \subfigure{
        \centering
    \includegraphics[width=0.48\textwidth]{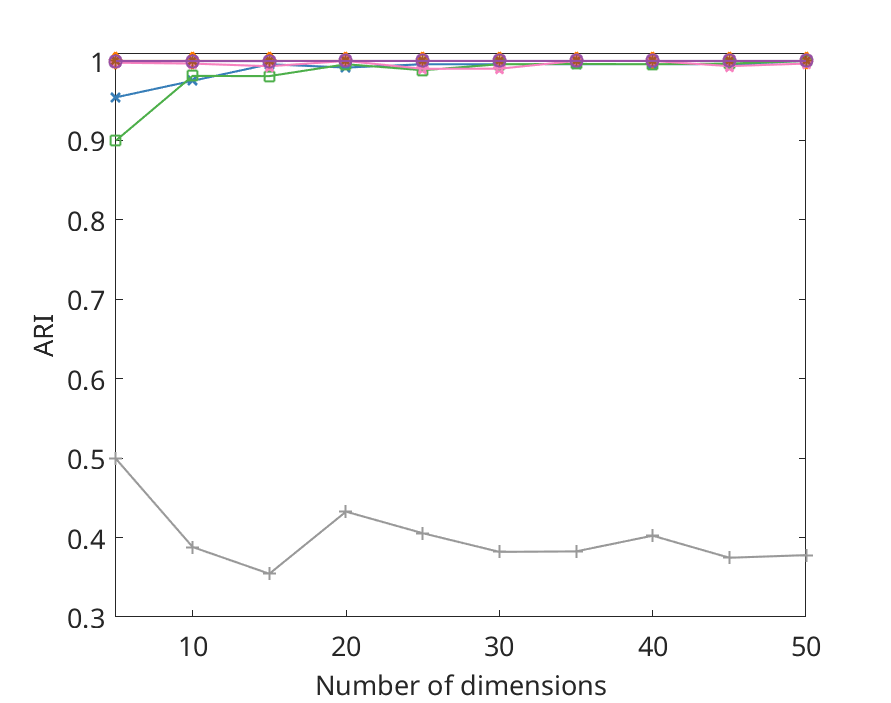} 
           }
           &
   \subfigure{
      \centering
       \includegraphics[width=0.48\textwidth]{ 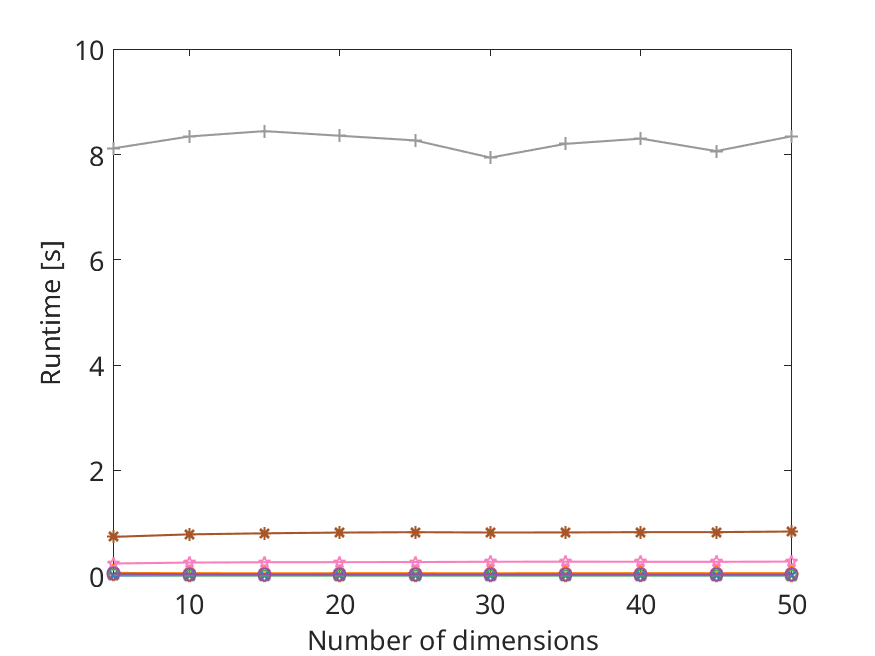} 
 }
  \\
     \subfigure{
        \centering
    \includegraphics[width=0.48\textwidth]{ 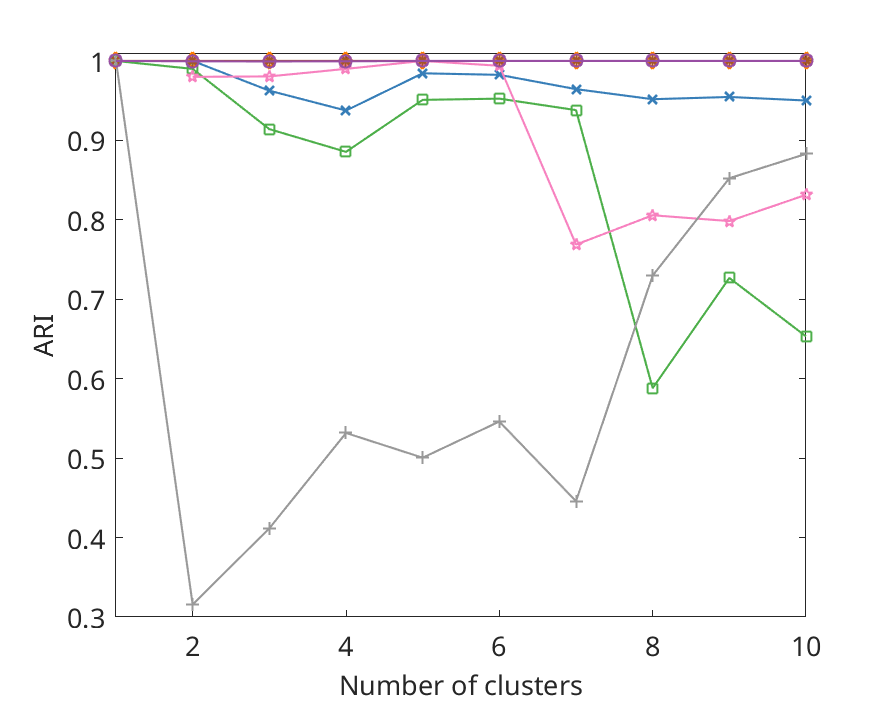} 
}
        &
     \subfigure{
        \centering
    \includegraphics[width=0.48\textwidth]{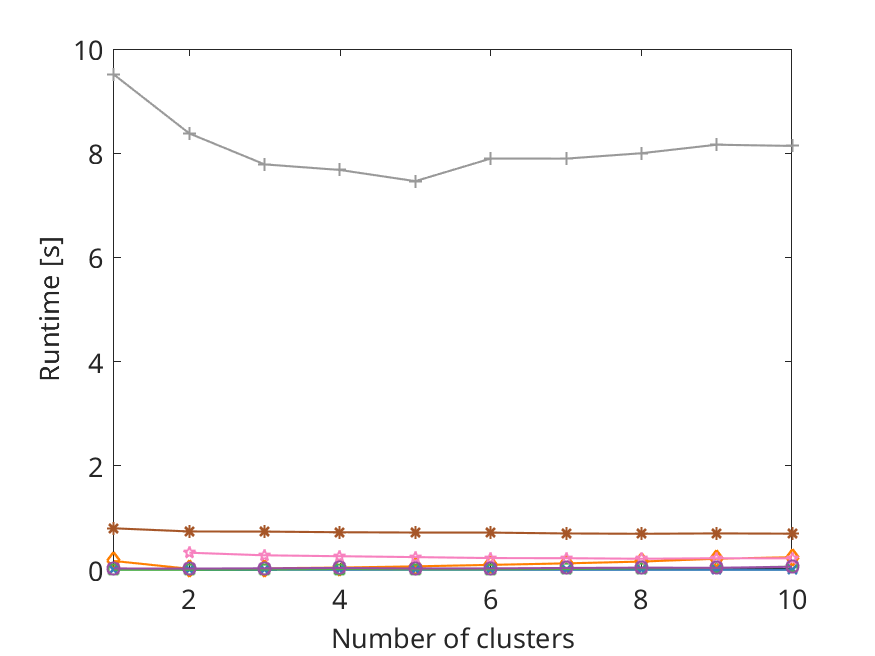} 
           }
    \end{tabular}
    \centering
        \includegraphics[width=0.875\textwidth]{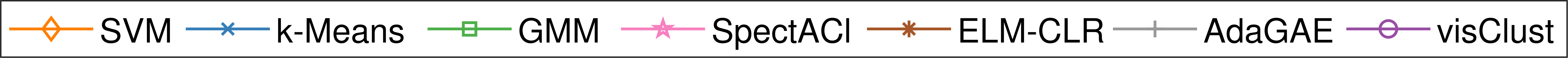} 
    \caption{\scriptsize 
    {\color{black}ARI (mean) and runtime (mean) of 100 independent runs for varying number of data points $m$ (top), dimensions $d$ (middle) and clusters $n_c$ (bottom) for the data set described in Section \ref{datarun}. SpectACl requires at least 2 clusters. Default: $m = 1000$, $d = 5$ and $n_c = 4$.} }
    \label{ariruntime}
\end{figure}

For the experiments provided in Figure \ref{ariruntime}, we use the function \textit{make\_blobs} from the module \py{sklearn} with standard deviation 0.05 as described in the Python documentation. 
The first cluster is created at the origin. The next cluster is centered exactly $2.5$ away from one of the existing clusters and at least $2.5$ from all other clusters. This procedure is repeated until the wanted number of clusters is reached.
When increasing the dimension of the data set, we draw the new data point entries from a uniform distribution ranging from the entry-wise minimum to the entry-wise maximum of all existing data points. 

\paragraph{Runtime experiments}\label{datarun}
In Figure \ref{ariruntime} we illustrate the runtime need of the selected algorithms, described in Section \ref{compalg}. The time is measured from starting the algorithm until termination, i.e. yielding an output partition. The initial setting of the above mentioned parameters is $m = 1000$, $d = 5$ and $n_c = 4$, whereas always one of the three parameters is varied. 

\paragraph{Storage}

\begin{figure}
\centering
    \includegraphics[width=0.7\textwidth]{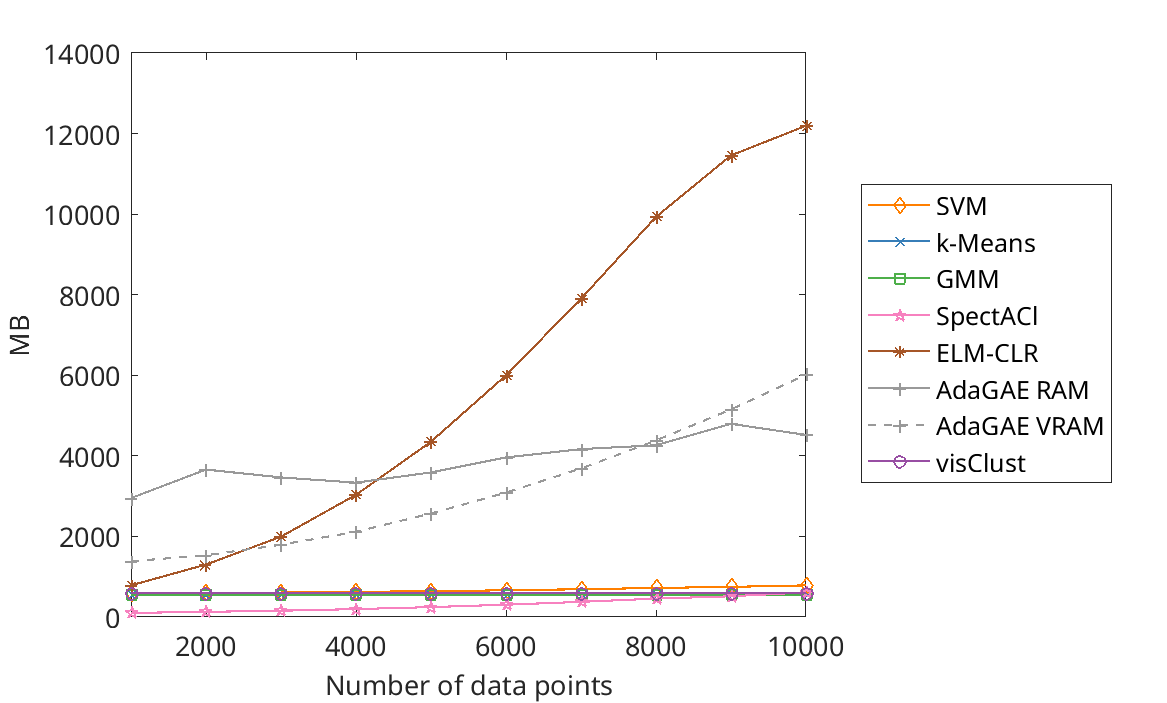} 
    \caption{\scriptsize\color{black}RAM (in MB) needed for the runtime experiments for a varying number of data points shown in Figure \ref{ariruntime} with synthetic data, where the number of data points $m$ varies, $d = 5$ and $n_c = 4$.  {\color{black} The \textit{AdaGAE} algorithm has been computed with GPU acceleration, therefore we report the needed RAM as well as Video RAM (VRAM).}
    }
    \label{storage}
\end{figure}

The RAM storage needed for the computations of the experiments {\color{black}for a varying number of data points in Figure \ref{ariruntime}} is stated in Figure \ref{storage}.

\paragraph{ARI}
{\color{black}In Figure \ref{ariruntime} we show the achieved ARI of the selected algorithms on the varying synthetic data sets used for the runtime experiments.}

\begin{figure}
\centering

   \includegraphics[width=0.49\textwidth]{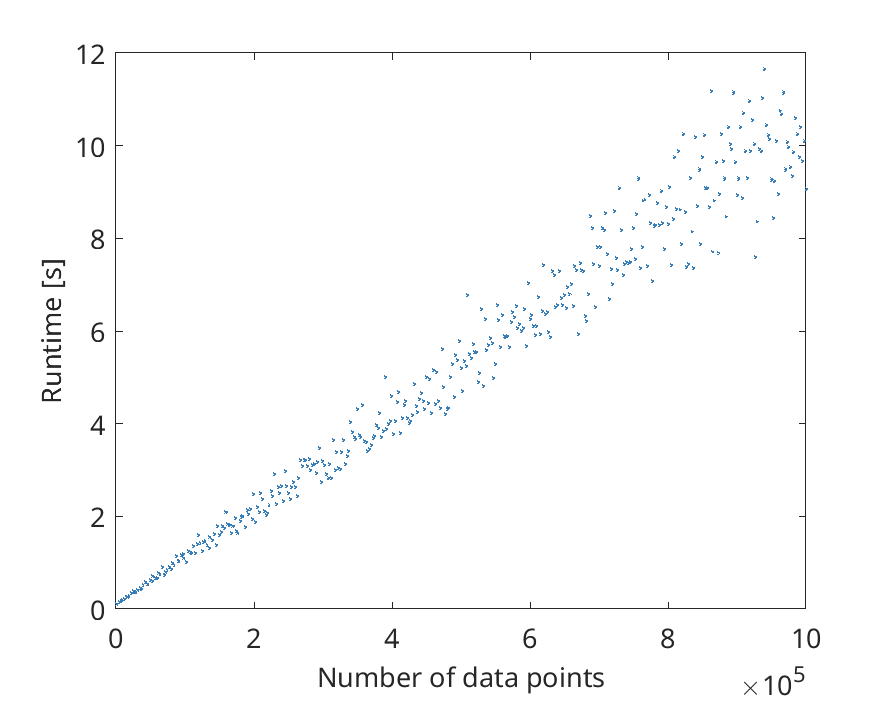}
        \includegraphics[width=0.49\textwidth]{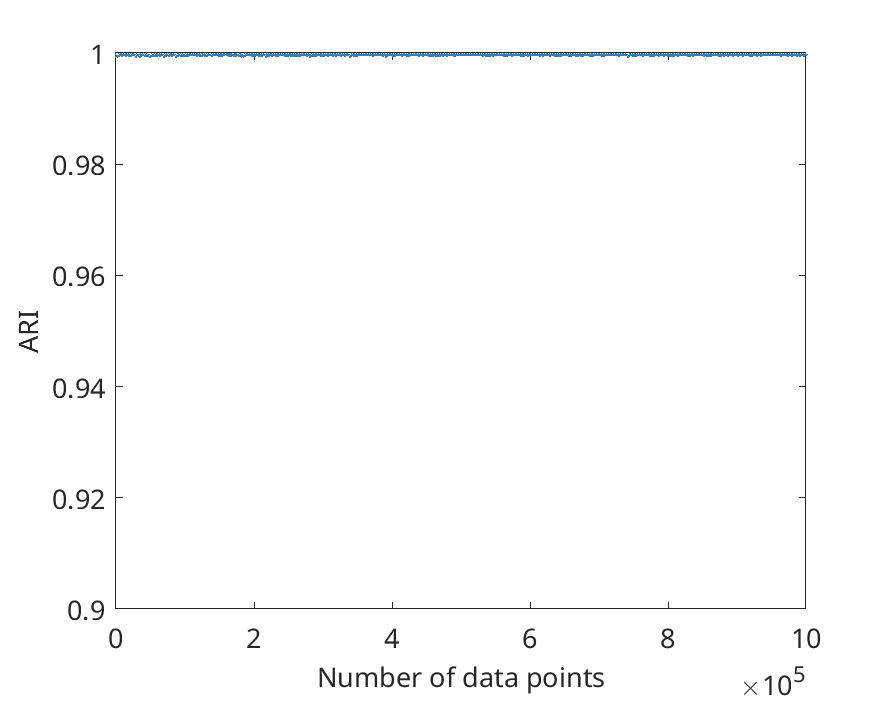} 

    \caption{\scriptsize Time (mean) of 100 independent runs in sec for termination of \textit{visClust} on synthetic data. {\color{black}The corresponding ARI (right) is constantly $\sim 1$ for all numbers of data points.}}
    \label{subsampfig}
\end{figure}

\paragraph{Runtime visClust}\label{runsub}
In the experiments shown in Figure \ref{subsampfig} we measure the time until termination of \textit{visClust} for the synthetic data set described in Section \ref{datas} for up to $1$ million data points {\color{black} and show the corresponding achieved ARI. }

\subsubsection{Publicly available test data sets} \label{publicdatasets}
\paragraph{Data sets} We use several different open source data sets with a wide range of properties regarding number of dimensions, data points and clusters. The data sets 'Iris', 'Wine', 'Seeds', 'Thyroid Disease' ('Thyroid'), 'Ecoli', 'Breast Cancer Wisconsin (Diagnostic)' ('Cancer'), 'Banknote Authentication' ('Banknotes'), 'Optical Recognition of Handwritten Digits' ('Digits', in the supplementary material), 'Wireless Indoor Localization' ('Wifi') and 'Mushroom' are used as provided at the UCI repository\footnote{\color{black}\url{http://archive.ics.uci.edu/ml}}. The dataset 'Yeast' consists of gene-expression data as attached in \cite{yeast}. The properties of all data sets can be found in Table \ref{tab:datasets}. For our experiments, all data sets have been scaled component-wise between $-1$ and $1$ using the \textit{mapminmax} function in Matlab. 

\begin{table}
\centering
\resizebox{0.8\columnwidth}{!}{
	\begin{tabular}{ |c||c|c|c|c|  }
		\hline
		& \textbf{Data points} & \textbf{Dimension} & \textbf{ Clusters}& \textbf{ Cluster division}\\
		\hline
		\hline
		Iris&150&4&4&25\%, 25\%, 25\%, 25\%\\
		\hline
		Wine&178&13&3&40\%, 33\%, 27\%\\
		\hline
		Yeast&205&80&4&45\%, 41\%, 7\%, 7\%\\
		\hline
		Seeds&210&7&3&33\%, 33\%, 33\%\\
		\hline
		Thyroid&215&5&3&70\%, 16\%, 14\%\\
		\hline
		Ecoli&336&7&8&\begin{tabular}{c}
			42.5\%, 22.9\%, 15.5\%,  10.4\%,\\  6\%, 1.5\%, 0.6\%, 0.6\%
		\end{tabular}\\
    \hline
Cancer&569&30&2&63\%, 37\%\\
    \hline
Banknotes&1372&4&2&55\%, 45\%\\
    \hline
Wifi&2000&7&4&25\%, 25\%, 25\%, 25\%\\
\hline
Mushroom&8124&22&2&51.7\%, 48.3\%\\
\hline
\end{tabular} }
\caption{\label{tab:datasets} \scriptsize{Properties of the publicly available test data sets.}}
\end{table}

\paragraph{ARI}
In Table \ref{tab:datasetsari} we state the ARI achieved by SVM and the unsupervised methods with default input parameters, when conducting $100$ independent runs of the methods applied to the chosen data sets. 
\begin{table}
\vspace{-0.29cm}
\centering
\resizebox{0.8\columnwidth}{!}{
	\begin{tabular}{ |c||c||c|c|c|c|c|c| }

		\hline
\textbf{Algorithm}& \textbf{SVM} & \textbf{k-Means} & \textbf{GMM}& \textbf{SpectACl}&\textbf{ELM-CLR} &\textbf{\color{black}AdaGAE}& \textbf{visClust}\\
		\hline
		\hline
		Iris&\makecell{$0.953 \pm 0.05$\\$0.883 \pm 0.11$} &\makecell{$0.844 \pm 0.11$\\$0.666 \pm 0.12$}&\makecell{$0.875 \pm 0.16$\\$0.770 \pm 0.21$}&\makecell{$0.688 \pm 0.01$\\$0.497 \pm 0.00$}&\makecell{$0.867 \pm 0.00$\\$0.677 \pm 0.00$} &\makecell{$0.799 \pm 0.11$\\$0.575 \pm 0.18$} & \makecell{$\mathbf{0.963} \pm 0.01$\\$\mathbf{0.895} \pm 0.02$} \\
		\hline
		Wine&\makecell{$0.977 \pm 0.03$\\$0.935 \pm 0.09$}&\makecell{$0.949 \pm 0.01$\\$0.849 \pm 0.02$}&\makecell{$0.757 \pm 0.26$\\$0.572 \pm 0.27$}&\makecell{$0.793 \pm 0.01$\\$0.493 \pm 0.01$}&\makecell{$\mathbf{0.970} \pm 0.00$\\$\mathbf{0.910} \pm 0.01$}&\makecell{$0.627 \pm 0.11$\\$0.270 \pm0.15 $}&\makecell{$0.883 \pm 0.04$\\$0.683 \pm 0.10$}\\
		\hline
		Yeast&\makecell{$0.848 \pm 0.08$\\$0.629 \pm 0.21 $}&\makecell{$\mathbf{0.966} \pm 0.04$\\$\mathbf{0.933} \pm 0.07$} &-&\makecell{$0.578 \pm 0.01$\\$0.193 \pm 0.01$}&\makecell{$0.966 \pm 0.00$\\$0.922\pm 0.00$}&\makecell{$0.619 \pm 0.09$\\$0.425 \pm 0.12$}&\makecell{$0.583 \pm 0.05$\\$0.375 \pm 0.06$}\\
		\hline
		Seeds&\makecell{$0.933 \pm 0.06$\\$0.810 \pm 0.16$}&\makecell{$0.889 \pm 0.00$\\$0.701 \pm 0.01$}&\makecell{$0.863 \pm 0.07$\\$0.668 \pm 0.11$}&\makecell{$0.634 \pm 0.01$\\$0.366 \pm 0.01$}&\makecell{$\mathbf{0.933} \pm 0.00$\\$\mathbf{0.813} \pm 0.00$}&\makecell{$0.679 \pm 0.11$\\$0.346 \pm 0.17$}&\makecell{$0.903 \pm 0.03$\\$0.739 \pm 0.07$}\\
		\hline
		Thyroid &\makecell{$0.949 \pm 0.05$\\$0.829 \pm 0.17$}&\makecell{$0.880 \pm 0.03$\\$0.626 \pm 0.06$}& -& \makecell{$0.814\pm 0.00$\\$0.656 \pm 0.00$}&\makecell{$0.767 \pm 0.00$\\$0.372 \pm 0.00$}&\makecell{$0.558 \pm 0.10$\\$0.167 \pm 0.12 $}&\makecell{$\mathbf{0.917} \pm 0.03$\\$\mathbf{0.797} \pm 0.08$}\\
		\hline
		Ecoli &\makecell{$0.869 \pm 0.08$\\$0.745 \pm 0.16$}&\makecell{$0.602 \pm 0.08$\\$0.440\pm 0.08$}&-&\makecell{$0.554 \pm 0.02$\\$0.407 \pm 0.03$}&\makecell{$0.603\pm 0.07$\\$0.468 \pm 0.08$}&\makecell{$0.520 \pm 0.05$\\$0.319 \pm 0.07 $}&\makecell{$\mathbf{0.676} \pm 0.08$\\$\mathbf{0.474} \pm 0.14$}\\
\hline
Cancer&\makecell{$0.965 \pm 0.02$\\$0.864 \pm 0.08$}&\makecell{$0.928 \pm 0.00$\\$0.732 \pm 0.00$}&-&\makecell{$0.796 \pm 0.01$\\$0.348 \pm 0.01$} &\makecell{$\mathbf{0.932} \pm 0.01$\\$\mathbf{0.748 }\pm 0.04$}&\makecell{$0.622\pm 0.10$\\$0.098 \pm 0.15$}&\makecell{$0.827 \pm 0.04$\\$0.432 \pm 0.10$}\\
\hline
Banknotes&\makecell{$1.000 \pm 0.00$\\$1.000 \pm 0.00$}&\makecell{$0.575 \pm 0.00$\\$0.022 \pm 0.00$}&\makecell{$0.586 \pm 0.14$\\$0.103 \pm 0.26$}&\makecell{$0.594 \pm 0.00$\\$0.035 \pm 0.00$}&\makecell{$\mathbf{0.992} \pm 0.00$\\$\mathbf{0.968 }\pm 0.00$}&\makecell{$0.601 \pm 0.08$\\$0.067 \pm 0.10$}& \makecell{$0.967 \pm 0.02$\\$0.875 \pm 0.06$} \\
\hline
Wifi&\makecell{$0.983 \pm 0.01$\\$0.956 \pm 0.02$}&\makecell{$\mathbf{0.908} \pm 0.08$\\$\mathbf{0.804} \pm 0.11$}&\makecell{$0.810 \pm 0.12$\\$0.679 \pm 0.19$}&\makecell{$0.686 \pm 0.00$\\$0.368 \pm 0.00$}&\makecell{$0.774 \pm 0.00$\\$0.656 \pm 0.00$}&\makecell{$0.443 \pm 0.07$\\$0.094 \pm0.09 $}&\makecell{$0.720 \pm 0.07$\\$0.625 \pm 0.13$}\\
\hline
Mushroom&\makecell{$1.000 \pm 0.00$\\$1.000 \pm 0.00$} &\makecell{$0.644 \pm 0.06$\\$0.098 \pm 0.08$}& - &\makecell{$0.650 \pm 0.02$\\$0.091 \pm 0.02$}& - & \makecell{$0.551 \pm 0.04$\\$0.018 \pm 0.02 $}&\makecell{$\mathbf{0.709} \pm 0.13$\\$\mathbf{0.243} \pm 0.20$} \\
\hline
\end{tabular} }
\caption{\label{tab:datasetsari} \scriptsize{{\color{black}ACC (mean $\pm$ standard deviation) and} ARI (mean $\pm$ standard deviation) of $100$ independent runs of the selected algorithms with input parameters in default setting evaluated on the publicly available data sets. GMM was excluded when the rate of failing in the default setting was more than $10\%$. {\color{black}ELM-CLR is not able to detect a partition with two clusters for the mushroom data set.} The best result achieved by an unsupervised methods is bold.}}
\end{table}

\begin{table}
\centering
\resizebox{0.9\columnwidth}{!}{
\begin{tabular}{ |c||c|c|c|c|c|c|c|c|c|c| }
\hline
\textbf{Data} & Iris & Wine & Yeast & Seeds & Thyroid & Ecoli & Cancer & Banknotes &Wifi&Mushroom \\ 
\hline
\hline
\textbf{visClust} default& \makecell{$\mathbf{0.963}$\\$\mathbf{0.895}$} & \makecell{$0.883$\\$0.683$} &\makecell{$\mathbf{0.583}$\\$\mathbf{0.375}$}& \makecell{$0.903$\\$0.739 $} & \makecell{$0.917$\\$0.797 $}  &\makecell{$0.676$\\$0.474$} & \makecell{$0.827$\\$0.432$} &\makecell{$0.967$\\$0.875$} &\makecell{$0.720$\\$0.625$}&\makecell{$0.709$\\$0.243$}\\
\hline
\textbf{visClust} optimal& \makecell{$\mathbf{0.963}$\\$\mathbf{0.895}$} & \makecell{$0.928$\\$0.794$} & \makecell{$\mathbf{0.916}$\\$\mathbf{0.804}$} & \makecell{$0.917$\\$0.772$} & \makecell{$0.926$\\$0.824$} &\makecell{$0.728$\\$0.589$} &  \makecell{$0.855$\\$0.509$} & \makecell{$0.970$\\$0.885$}  & \makecell{$0.814$\\$0.735$}& \makecell{$0.728$\\$0.245$}\\
\hline
\end{tabular} }
\caption{\label{tab:datasetsari2} \scriptsize{{\color{black}ACC (mean) and }ARI (mean) of 100 independent runs using \textit{visClust} with input parameters in default setting versus optimal input parameters computed regarding the target data. Biggest and smallest changes are bold.}}
\end{table}
\paragraph{Optional parameters visClust} 
So far, the evaluation of \textit{visClust} was conducted using the default input parameters (see Section \ref{inpar}). To investigate if the results of \textit{visClust} can be increased when using tailored input parameters, we iterate over chosen parameter ranges regarding the threshold $t$, the scaling parameter $s$ and vary between the correct and the default cluster division (uniform). The optimal parameters were chosen for $s$ in the range from $0.5$ to $2$ and for the threshold $t$ in the range from $0.01$ to $0.3$. The results in the default and optimal setting are stated in Table \ref{tab:datasetsari2}.

\begin{figure}
\centering
    \includegraphics[width=0.8\textwidth]{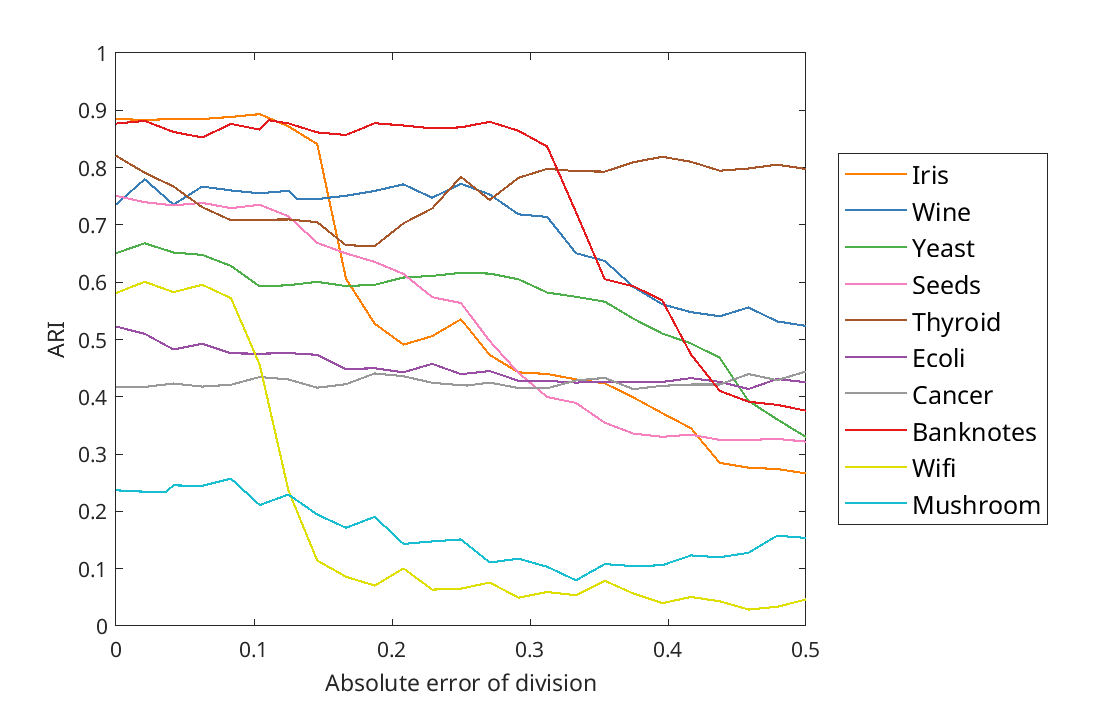} 
    \caption{\scriptsize ARI (mean) of $100$ independent runs of provided partitions by \textit{visClust} when using a varying deviation of the correct cluster division as input parameter. (Left) correct division, (Right) $50\%$ deviation. 
    }
    \label{devdiv}
\end{figure}

Moreover, we analyzed the dependence of the success of the algorithm regarding deviation from the correct cluster division. In Figure \ref{devdiv} the ARI of the resulting partitions are stated when deviating the division from the correct one up to $50\%$. All other input parameters are set to the default setting. 

\section{Discussion}\label{disc}
{\color{black}The visual results in Figure \ref{visComp} show that \textit{visClust} performs well in all the classification challenges. As expected, \textit{k-Means} and \textit{GMM} struggle on the intertwined data structures in rows 1 and 2, \textit{spectACl} {\color{black}and \textit{AdaGAE}} face challenges with most tasks. When asked to partition the cluster wrongly into $3$ clusters (last row), \textit{visClust} is the only unsupervised method that does not return a wrong partition. It acts superior to the other compared unsupervised classification methods failing in at least one of the toy examples. }

In the runtime experiments with synthetic data, cf. Figure \ref{ariruntime}, we observe that the varying number of data points, dimensions and clusters have nearly no impact on the runtime of \textit{visClust}. The algorithm needs less than a second to terminate, being marginally slower than \textit{k-Means} and \textit{GMM}. 
{\color{black}On the other hand, it is faster than \textit{ELM-CLR} and \textit{spectACl}. In particular, \textit{spectACl} behaves linearly and the runtime of \textit{ELM-CLR} {\color{black}and \textit{AdaGAE}} increase exponentially with the number of data points leading to a very high runtime already for $10^4$ data points. 
The corresponding ARI to the simple runtime experiments is high for most algorithms and all data variations. However, we observe that \textit{GMM} and \textit{k-Means} yield lower results in comparison to the other methods, and, like \textit{spectACl}, behave unstable when it comes to varying number of clusters.} {\color{black}\textit{AdaGAE} shows very low numbers for the ARI, its performance increases for a higher amount of data, which probably is due the design for bigger data sets.} Here, \textit{visClust} is the only method that demonstrates outstanding behavior regarding both, runtime and ARI, in these experiments.

Moreover, we analyze the needed RAM for the runtime experiments, see Figure \ref{storage}, and observe that most methods appear to be (nearly) constant in their RAM requirement, {\color{black}\textit{spectACl} behaves linearly again}. However, for \textit{ELM-CLR} {\color{black} and \textit{AdaGAE}} it depends highly on the size of the data set; for $10^4$ data points \textit{ELM-CLR} needs already $\sim$12.2GB and {\color{black} \textit{AdaGAE} $\sim$10GB in total. In the case of \textit{AdaGAE}, 6GB video RAM is used, which restricts the applicability since this is not necessarily available on a standard computer.} Moreover, it increases exponentially and therefore it is not feasible to use the methods on bigger data sets on a conventional PC. In particular, here, \textit{ELM-CLR} completely fails after 47k input data points. 
In more experiments regarding runtime, cf. Figure \ref{subsampfig}, we observe - as expected theoretically - that the runtime of \textit{visClust} grows linearly in dependence of the number of data points. We see that the method terminates fast even for $1$ million data points ($\sim$10 seconds). 

With the results of the open-source data sets clustered in the default setting of all methods, cf. Table \ref{tab:datasetsari}, the observation is confirmed that \textit{ELM-CLR} and \textit{visClust} outperform \textit{k-Means}, \textit{GMM}, \textit{spectACl} and {\color{black}\textit{AdaGAE}} on most data sets regarding the ARI. The \textit{GMM} was excluded when the processing failed in more than $10\%$ of the runs. \textit{visClust} and \textit{ELM-CLR} show in this experiment superior behavior. \textit{ELM-CLR} asks in comparison for the most obligatory input parameters, cf. Table \ref{paramalg}. {\color{black}For one of these $7$ parameters a default value for UCI data sets is provided in the corresponding paper \cite{LIU2018126}. Most of the parameters are not obvious to choose when downloading the code without further optimization using target data. For the experiments in this paper, we chose the input parameters as in a code example from the authors of the paper.} In comparison, \textit{visClust} only asks for one standard input parameter, the number of clusters. All other parameters are set to the default suggestions that are provided directly in the code. 

In order to evaluate the optional parameters in more detail, we tested \textit{visClust} regarding its stability using a wrong cluster division as input versus using the correct division of clusters. First, we checked its stability in the default setting, when uniform cluster sizes are used as input parameters even though the cluster sizes might strongly deviate amongst each other. In Figure \ref{devdiv} we observe that for up to 
$20\%$ deviation of the correct division most data sets do not show decreased performance regarding the ARI. For around half of the data sets the ARI drops at the $30\%$ deviation mark. For a third of the data sets the ARI does not decrease significantly even for $50\%$ deviation. These results are very promising since one would expect that the user does have some information if the expected cluster sizes are very unbalanced. In Table \ref{tab:datasetsari2} results of \textit{visClust} on the same data sets as evaluated in Table \ref{tab:datasetsari} are stated, where this time all input parameters have been optimized. It reveals that in some cases the results can be improved significantly by choosing optimal input parameters, whereas in most cases it yields minor improvement. This is again very promising, since we do not expect the average user to spend lots of time on parameter optimization. But if so, this extra effort probably will show improvement to the default setting. 

\section{Limitations}
{\color{black}First, it has to be mentioned that the algorithm asks for one obligatory input parameter in the default setting, the number of clusters. Most standard clustering algorithms, including the ones we compared to, ask for that parameter, aiming to find the best partition in regards to the provided number of clusters. Nonetheless, to overcome this limitation, there are many enhancements and novel methods that do not ask for that input parameter and therefore act beneficial when there is no knowledge about the number of clusters. This limitation can be overcome by applying estimators or optimization of the number of clusters beforehand, see e.g.~\cite{LIANG20122251, KOLESNIKOV2015941,UNLU201933}. The implementation of \textit{visClust} also provides the option to run the method without any input parameter. In that case the clustering is determined by choosing the numbers of clusters that appear in the lower dimensional representations most frequently. 

The algorithm is based on linear orthogonal projections mapping the data to a lower dimensional space, distorting the structural information within each data point when the dimension of the original data is bigger than $3$. This enables varied lower dimensional representations but has the drawback to not being able to sufficiently apply the algorithm to imaging data, which is naturally high-dimensional and depends on its structural information. In the Supplementary Material we show an experiment with a digit image data set. To overcome this limitation, we allow the user to choose the non-linear projection \textit{t-SNE} \cite{tsne} rather than orthogonal projections as an additional non-obligatory parameter.  

Due to computational efficiency, the algorithm is limited to $k=2$ and $k=3$, which leads to challenges for data sets that do not decompose nicely in these dimensions. To tackle that problem, we implemented an alternative route, {\color{black}described in the Supplementary Material}, that identifies the clusters separately in turn. For the tested data sets this works sufficiently well, but nonetheless, data sets consisting of important high-dimensional information may not be well described by a representation in two or three dimensions. In these cases \textit{visClust} will not be a good choice of method. 

Lastly, we want to shortly discuss that evaluating clustering algorithms by conducting classification tasks is limiting. The standard problem of classification aims to partition provided data into specific, humanly chosen, classes. Clustering as such is looking for meaningful partitions within a data set that may not correspond to a-priory chosen classes, but may even provide insightful information of the data, discovering similarity properties within the data. When testing clustering algorithms with classification data, it could happen, that the algorithm would indeed identify some meaningful classes, but not the ones that were chosen for the classification, and therefore the accuracy results in a low number. As some kind of quantitative comparison is needed, and it is the state-of-the-art to provide such experiments, we conducted this kind of experiments but also provide some supplementary experiments with synthetic data. Nonetheless, a wider approach to evaluate the clustering method would be beneficial.

\section{Conclusion}
We proposed a novel clustering algorithm, called \textit{visClust}, that is based on visual interpretation of lower dimensional representations obtained by projections from the Stiefel manifold. The algorithm is well understood, stable and fast regarding a varying number of data points, dimensions and clusters. We showed in a comprehensive amount of experiments that the algorithm in the default setting yields good results concerning ACC, ARI, runtime and needed RAM. 

In experiments with standard synthetic data it becomes obvious that \textit{visClust} clearly outperforms standard choices such as \textit{k-Means} or \textit{GMM} for data that is more complicated, e.g. intertwined. Also on publicly available test data sets \textit{visClust} acted superior to \textit{k-Means} and \textit{GMM}. The density based method \textit{spectACl} {\color{black}as well as the deep clustering method \textit{AdaGAE} are both outperformed by \textit{visClust} in the default setting in all experiments due to the need of input parameter optimization.}

The adaptive graph based method \textit{ELM-CLR} and \textit{visClust} outperform each other regarding ARI for different data sets. In more detailed analyses we observe that \textit{ELM-CLR} has the huge drawback of being unfeasible for larger data sets, due to its immense storage need and high runtime, as well as not providing clear suggestions for the wide range of obligatory input parameters. For \textit{visClust} just one obligatory input parameter is needed in the default setting and clear suggestions for choosing the additional optional input parameters are provided. These parameters allow improvement while not being necessary for a sufficient clustering result. 

In summary, \textit{visClust} is a fast clustering method and straightforward to use, yielding good results in default setting even for more complicated data sets. Therefore, we believe that \textit{visClust} is an important contribution to the landscape of clustering methods, overcoming several limitations of other proposed algorithms. The code is made available on GitHub, cf. {\footnote{\color{black}\url{https://github.com/charmed-univie/visclust}}. }

\section*{Acknowledgement}
This work was funded by the Vienna Science and Technology Fund (WWTF) through project VRG12-009, the Austrian Science Fund (FWF) through project T-1307 and project P33217 as well as the ÖAW/Unesco/L'ORÉAL Austria through the fellowship 'FOR WOMEN IN SCIENCE'.

\bibliographystyle{elsarticle-num}
{\scriptsize \bibliography{references.bib}}
\newpage
\section{Appendix}
\subsection{Implementation details}

\begin{itemize}

\item The algorithm starts with $k = 2$ and if it does not terminate it swaps to $k = 3$. We observed that surprisingly for most data sets $k =2$ already yields a sufficient solution and $k = 3$ never improved the clustering outcome crucially in our experiments. Since the computational cost and time increases with the dimension $k$ we restricted the algorithm to $k=2$ and $k=3$.

\item As a trade-off between run-time and sufficient covering we use $5000$ projections in $\G_{2,d}$ and $2000$ projections in $\G_{3,d}$. 

\item In case of the need for even faster performance, e.g. when using huge data sets, we allow subsampling by random choice of a subset from the full data set. The remaining points are assigned eventually to the optimal cluster using the k-nearest neighbors algorithm, cf. \cite{10.1145/355744.355745}.

\item The scaling factor $s$, weighting the standard deviation of the Gaussian filter, is changed every $250$ iterations in the following manner: if after $250$ iterations more than $80\%$ of the found partitions are not containing enough clusters, $s$ is increased by $25\%$. If on the other hand $80\%$ of the found partitions contain too many clusters, $s$ is decreased by $25\%$.

\item Outliers are discarded at $4$ standard deviations. These points are assigned to a partition in the last step of the algorithm by finding the nearest cluster in Euclidean distance. 

\item If the desired number of clusters are not identified, we switch to a recursive binary clustering. We start by aiming to identify the supposedly smallest cluster versus all other potential clusters. Once identified, we proceed with binary clustering on the remaining data until we reach the desired number of clusters.

\end{itemize}

\subsection{Evaluation metrics}
For the evaluation of our classification experiments we use the performance metrics accuracy (ACC) and an adjusted Rand index (ARI) with a one-sided permutation model. 
\subsection{Accuracy (ACC)}
The accuracy of a classification problem corresponds to the number of correctly classified data points in relation to the whole data set. With $m$ data points $x = \{x_i\}_{i=1}^m$ and corresponding target classification $c \in \{1, \dots, n_c\}^m$ into $n_c$ classes it is given by the fraction
\[ ACC(c,\tilde{c})= \frac{\sum_{i=1}^m \delta_{c_i,\tilde{c}_i}}{m}
\]
for the output clustering $\tilde{c} \in \{1, \dots, n_c\}^m$, where $\delta$ denotes the Kronecker delta function. 
The optimal permutation of all clusters versus classes is then computed and used for the evaluation. 
\subsection{Adjusted Rand index}
\begin{table}
\begin{center}
\begin{tabular}{ c |c c c c|c }
 $A/B$ & $B_1$ & $B_2$ & $\ldots$ & $B_{n_{cB}}$ &Sums\\ 
 \hline
 $A_1$ & $n_{1,1}$ & $n_{1,2}$ &$\cdots$&$n_{1,n_{c}}$&$a_1$\\  
 $A_2$ & $n_{2,1}$ & $n_{2,2}$ &$\cdots$&$n_{2,n_{c}}$& $a_2$\\
  $\vdots$ & $\vdots$  & $\vdots$  &$\ddots$&$\vdots$ &$\vdots$\\  
 $A_{n_{c}}$ & $n_{n_{c},1}$ & $n_{n_{c},2}$ &$\cdots$&$n_{n_{c},n_{c}}$&$a_{n_{c}}$\\
 \hline
  Sums &$b_1$ & $b_2$ &$\cdots$&$b_{n_{c}}$&$\sum_{ij}n_{i,j}=m$\\
\end{tabular}
\end{center}
\caption{\scriptsize The contingency table for two partitions $A=\{A_1,\ldots,A_{n_{c}}\}$ and $B=\{B_1, \ldots,B_{n_{c}}\}$, where $n_{i,j}=|A_i \cap B_j|$, is the number of elements in the intersection of the clusters $A_i\in A$ and $B_j\in B$ as presented in \cite{arinew}.}
\label{contingencytable}
\end{table}
We follow the definition of the adjusted Rand index for a one-sided permutation model and a fixed number of clusters as in \cite[p.~3]{arinew}. For two partitions $A=\{A_1,\ldots,A_{n_{c}}\}$ and $B=\{B_1, \ldots,B_{n_{c}}\}$ it is given by
\begin{align*}
    \mathrm{ARI}(A,B)&=\frac{\mathrm{RI}(A,B)-\mathbb{E}[\mathrm{RI}(A,B)]}{1-\mathbb{E}[\mathrm{RI}(A,B)]},
\end{align*}
where
\begin{align*}
    \mathbb{E}[\mathrm{RI}(A,B)]&=\left(\frac{\mathrm{S}(m-1,n_{cA})}{\mathrm{S}(m,n_{cA})}\frac{\sum_j\binom{b_j}{2}}{\binom{m}{2}}\right)\\
    &\quad+\left(1-\frac{\mathrm{S}(m-1,n_{cA})}{\mathrm{S}(m,n_{cA})}\right)\left(1-\frac{\sum_j\binom{b_j}{2}}{\binom{m}{2}}\right),
\end{align*}
and $S(n,k)$ denote the Stirling numbers of second kind. They are given by
\begin{equation*}
    S(n,k)=\frac{1}{k!}\sum_{j=0}^k(-1)^{k-j}\binom{k}{j}j^n
\end{equation*}
for $n,k \in \N$. The Rand index $\mathrm{RI}(A,B)$ is defined by
\begin{equation*}
    \mathrm{RI}(A,B)=\frac{2\sum_{k=1}^{n_{c}}\sum_{m=1}^{n_{c}}\binom{n_{k,m}}{2}-\sum_{k=1}^{n_{c}}\binom{a_k}{2}-\sum_{m=1}^{n_{c}}\big(\binom{b_m}{2}+\binom{m}{2}\big)}{\binom{m}{2}}
\end{equation*}
for $a_k,b_m$ and $n_{k,m}$ as given in Table \ref{contingencytable}.
\subsection{t-Distributed Stochastic Neighbor Embedding (t-SNE)}
The t-Distributed Stochastic Neighbor Embedding (t-SNE) \cite{tsne} is a statistical method which transforms high-dimensional data sets non-linearly to lower dimensions while preserving some of the cluster structure. It is based on the Stochastic Neighbor Embedding method \cite{sne}. For our experiments, we used the official Matlab implementation \textit{tsne} from the \textit{Statistics and Machine Learning} toolbox.

\subsection{visClust on images with handwritten digits}

In this section, we study the performance of visClust using the `Optical Recognition of Handwritten Digits' (`Digits') dataset as provided at the UCI repository\footnote{\url{http://archive.ics.uci.edu/ml}} as well as in combination with the mentioned `t-SNE' algorithm. 

As can be seen in Table \ref{tab:datasetsari}, visClust fails to classify well when used on the original data, but clearly outperforms all other algorithms when using \textit{t-SNE} as pre-processing. A grid search over the recommended parameter range yielded no improvement of the clustering performance as presented in Table \ref{tab:gridsearchdigits}. Additionally, Figure \ref{wrongdistrdigits} highlights that visClust is stable up to an absolute error of the input division of approximately 20 percent. Therefore, we would recommend to test the 't-SNE` option when using \textit{visClust} for imaging data. 
\begin{table}
\centering
\resizebox{0.8\columnwidth}{!}{
	\begin{tabular}{ |c||c|c|c|c|  }
 \hline 
 		& \textbf{Data points} & \textbf{Dimension} & \textbf{ Clusters}& \textbf{ Cluster division}\\
\hline \hline
Digits&1797&64&10& \makecell{10.2\%, 10.1\%, 10.1\%, 10.1\%, 10.1\%,\\ 10\%, 10\%, 9.9\%, 9.8\%, 9.7\%}\\
    \hline
\end{tabular} }
\caption{\label{tab:datasets} \scriptsize{Properties of the data set digits.}}
\end{table}

\begin{table}
\vspace{-0.29cm}
\centering
\resizebox{0.8\columnwidth}{!}{
	\begin{tabular}{ |c||c||c|c|c|c|c|c| }

		\hline
\textbf{Algorithm}& \textbf{SVM} & \textbf{k-Means} & \textbf{GMM}& \textbf{SpectACl}&\textbf{ELM-CLR} & \textbf{AdaGAE}&\textbf{visClust}\\
		\hline
		\hline
Digits&\makecell{$0.491 \pm 0.09$\\$-1.064 \pm 0.54$}&\makecell{$\mathbf{0.730} \pm 0.06$\\$\mathbf{0.592} \pm 0.07$}&-&\makecell{$0.578 \pm 0.01$\\$0.193 \pm 0.01$}&\makecell{$0.688 \pm 0.02$\\$0.453 \pm 0.04$}&\makecell{$0.389 \pm 0.04 $\\$0.138 \pm 0.07$}&\makecell{$0.146 \pm 0.07 $\\$-0.137 \pm 0.23$}\\
\hline
Digits t-SNE&\makecell{$0.534 \pm 0.08$\\$-0.796 \pm 0.48$} &\makecell{$0.867 \pm 0.06$\\$0.807 \pm 0.06$}&\makecell{$0.817 \pm 0.07$\\$0.743 \pm 0.07$}&\makecell{$0.745 \pm 0.17$\\$0.580 \pm 0.27$}&\makecell{$0.688 \pm 0.02$\\$0.453 \pm 0.04$}& \makecell{$0.786 \pm 0.06$\\$0.679 \pm 0.08$}&\makecell{$\mathbf{0.948} \pm 0.01$\\$\mathbf{0.894} \pm 0.01$} \\
\hline
\end{tabular} }
\caption{\label{tab:datasetsari} \scriptsize{ACC (mean $\pm$ standard deviation) and ARI (mean $\pm$ standard deviation) of $100$ independent runs of the selected algorithms with input parameters in default setting evaluated on the publicly available data sets. GMM was excluded when the rate of failing was more than $10\%$. The best result achieved by an unsupervised methods is bold.}}
\end{table}
\begin{table}
\centering
\resizebox{0.4\columnwidth}{!}{%
\begin{tabular}{ |c||c|c| }
\hline
\textbf{Data} & Digits &Digits t-SNE\\ 
\hline
\hline
\textbf{visClust} default& \makecell{$0.146 $\\$-0.137$} &\makecell{$0.948$\\$0.894$}\\
\hline
\textbf{visClust} optimal&\makecell{$0.146 $\\$-0.137$} &\makecell{$0.948$\\$0.894$}\\
\hline
\end{tabular} }
\caption{\label{tab:gridsearchdigits} \scriptsize{ACC (mean) and ARI (mean) of 100 independent runs using \textit{visClust} with input parameters in default setting versus \textit{visClust}.}}
\end{table}

\begin{figure}
\centering
    \includegraphics[width=0.8\textwidth]{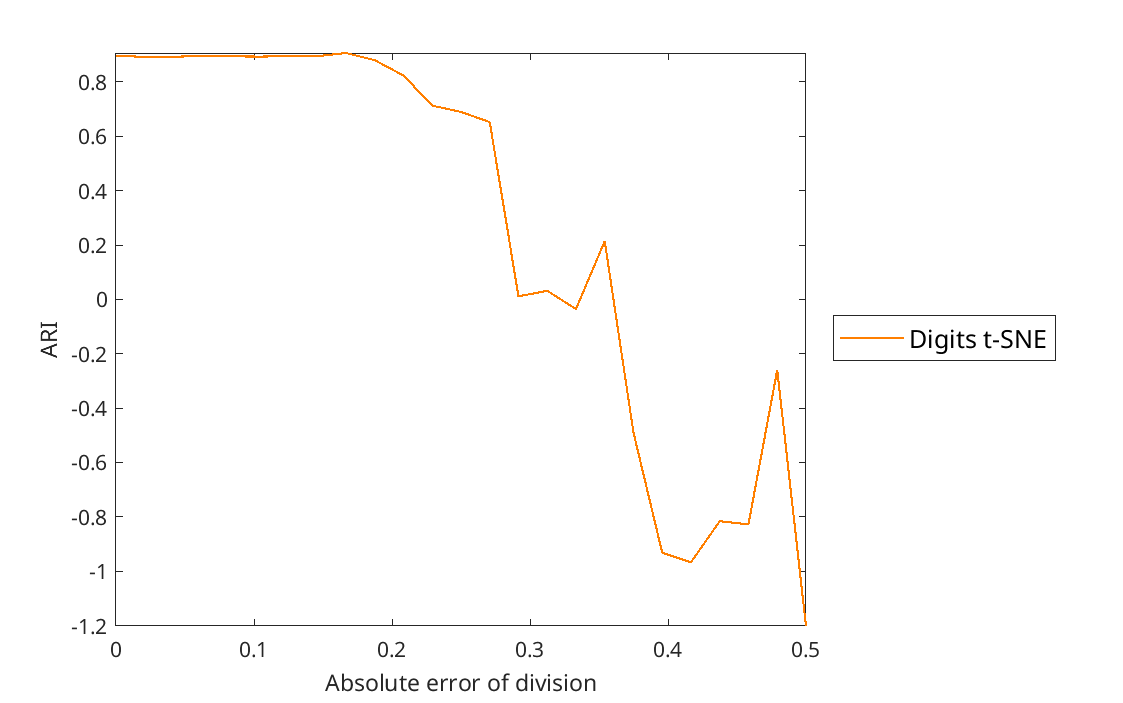} 
    \caption{\scriptsize ARI (mean) of $100$ independent runs of provided partitions by \textit{visClust} when using a varying deviation of the correct cluster division as input parameter. (Left) correct division, (Right) $50\%$ deviation.
    }
    \label{wrongdistrdigits}
\end{figure}

\subsection{Computation sample variance (Eq. 6, Section 2.2)}
\begin{align*}
\frac{1}{m-1} \sum_{i=1}^m \|x_i-\bar{x}\|^2 & = \frac{1}{m-1} \sum_{i=1}^m \left(\|x_i\|^2+\|\bar{x}\|^2-2\langle x_i,\bar{x}\rangle\right)\\
& = \frac{1}{m-1} \sum_{i=1}^m \|x_i\|^2+ \frac{1}{m-1} \sum_{i=1}^m \frac{1}{m^2}\sum_{j,k} \langle x_j,x_k\rangle - \frac{2}{m-1}\sum_{i} \frac{1}{m}\sum_{j} \langle x_i,x_j\rangle\\
& = \frac{1}{m-1} \sum_{i=1}^m \|x_i\|^2+ \frac{1}{m-1}\frac{1}{m}\sum_{j,k} \langle x_j,x_k\rangle - \frac{2}{m-1}\cdot\frac{1}{m}\sum_{i,j}\langle x_i,x_j\rangle\\
&= \frac{1}{m-1} \sum_{i=1}^m \|x_i\|^2-\frac{1}{m-1}\frac{1}{m}\sum_{i,j}\langle x_i,x_j\rangle\\
& = \frac{1}{m-1}\cdot\frac{1}{2}\left(\sum_{i}\|x_i\|^2+ \sum_{j} \|x_j\|^2 - \frac{2}{m}\sum_{i,j}\langle x_i,x_j\rangle\right)\\
& = \frac{1}{m-1}\cdot\frac{1}{2}\cdot\frac{1}{m}\left(\sum_{i,j} \|x_i\|^2+ \sum_{i,j}\|x_j\|^2 - 2\sum_{i,j}\langle x_i,x_j\rangle\right)\\
& = \frac{1}{m-1}\cdot\frac{1}{2}\cdot\frac{1}{m}\sum_{i,j}\left( \|x_i\|^2+ \|x_j\|^2 - 2 \langle x_i,x_j\rangle\right)\\
& = \frac{1}{m-1}\cdot\frac{1}{2}\cdot\frac{1}{m}\sum_{i,j}\|x_i-x_j\|^2\\
& = \frac{1}{m-1}\cdot\frac{1}{m}\sum_{i<j}\|x_i-x_j\|^2.
\end{align*}

\end{document}